\documentclass[10pt,logo,copyright]{nvidiatechreport}
\usepackage[authoryear,sort&compress,round]{natbib}

\usepackage[utf8]{inputenc} 
\usepackage[T1]{fontenc}    

\usepackage{parskip}        
\usepackage{url}            
\usepackage{booktabs}       
\usepackage{amsfonts}       
\usepackage{nicefrac}       
\usepackage{microtype}      
\usepackage{xcolor}         
\usepackage[dvipsnames]{xcolor} 
\usepackage{graphicx}
\usepackage{animate}        
\usepackage{subcaption}
\usepackage{tabularx}
\usepackage{makecell}
\usepackage{adjustbox}
\usepackage{setspace}
\newcolumntype{M}[1]{>{\centering\arraybackslash}m{#1}}
\usepackage{float}
\usepackage{tikz}
\usetikzlibrary{positioning,shapes,arrows}
\usepackage{amsmath,amsfonts,bm, bbm,leftindex}
\usepackage{multirow}
\usepackage{comment}
\usepackage{gensymb}
\usepackage{lipsum}
\usetikzlibrary{arrows.meta, positioning, fit}
\usepackage[para]{threeparttable}
\usepackage{tikz}
\usetikzlibrary{tikzmark}

\def\eqref#1{equation~\ref{#1}}

\def\1{\bm{1}}

\DeclareMathAlphabet{\mathsfit}{\encodingdefault}{\sfdefault}{m}{sl}
\SetMathAlphabet{\mathsfit}{bold}{\encodingdefault}{\sfdefault}{bx}{n}

\makeatletter
\let\save@mathaccent\mathaccent
\newcommand*\if@single[3]{%
  \setbox0\hbox{${\mathaccent"0362{#1}}^H$}%
  \setbox2\hbox{${\mathaccent"0362{\kern0pt#1}}^H$}%
  \ifdim\ht0=\ht2 #3\else #2\fi
  }
\newcommand*\rel@kern[1]{\kern#1\dimexpr\macc@kerna}
\newcommand*\widebar[1]{\@ifnextchar^{{\wide@bar{#1}{0}}}{\wide@bar{#1}{1}}}
\newcommand*\wide@bar[2]{\if@single{#1}{\wide@bar@{#1}{#2}{1}}{\wide@bar@{#1}{#2}{2}}}
\newcommand*\wide@bar@[3]{%
  \begingroup
  \def\mathaccent##1##2{%
    \let\mathaccent\save@mathaccent
    \if#32 \let\macc@nucleus\first@char \fi
    \setbox\z@\hbox{$\macc@style{\macc@nucleus}_{}$}%
    \setbox\tw@\hbox{$\macc@style{\macc@nucleus}{}_{}$}%
    \dimen@\wd\tw@
    \advance\dimen@-\wd\z@
    \divide\dimen@ 3
    \@tempdima\wd\tw@
    \advance\@tempdima-\scriptspace
    \divide\@tempdima 10
    \advance\dimen@-\@tempdima
    \ifdim\dimen@>\z@ \dimen@0pt\fi
    \rel@kern{0.6}\kern-\dimen@
    \if#31
      \overline{\rel@kern{-0.6}\kern\dimen@\macc@nucleus\rel@kern{0.4}\kern\dimen@}%
      \advance\dimen@0.4\dimexpr\macc@kerna
      \let\final@kern#2%
      \ifdim\dimen@<\z@ \let\final@kern1\fi
      \if\final@kern1 \kern-\dimen@\fi
    \else
      \overline{\rel@kern{-0.6}\kern\dimen@#1}%
    \fi
  }%
  \macc@depth\@ne
  \let\math@bgroup\@empty \let\math@egroup\macc@set@skewchar
  \mathsurround\z@ \frozen@everymath{\mathgroup\macc@group\relax}%
  \macc@set@skewchar\relax
  \let\mathaccentV\macc@nested@a
  \if#31
    \macc@nested@a\relax111{#1}%
  \else
    \def\gobble@till@marker##1\endmarker{}%
    \futurelet\first@char\gobble@till@marker#1\endmarker
    \ifcat\noexpand\first@char A\else
      \def\first@char{}%
    \fi
    \macc@nested@a\relax111{\first@char}%
  \fi
  \endgroup
}
\makeatother

\usepackage{times}
\usepackage{etoc}

\usepackage{algpseudocode}
\usepackage{algorithm}
\usepackage[algo2e]{algorithm2e}
\usepackage{multirow}
\usepackage[normalem]{ulem}
\usepackage{hyperref}       
\usepackage{url}            
\usepackage{booktabs}       
\usepackage{amsfonts}       
\usepackage{nicefrac}       
\usepackage{microtype}      
\usepackage{amsmath,amssymb}
\usepackage{gensymb}
\usepackage{algorithm}
\usepackage[algo2e]{algorithm2e}
\usepackage{color}
\usepackage{graphicx}
\usepackage{tabulary,multirow,overpic,xcolor}
\usepackage{wrapfig}
\usepackage{sidecap}
\usepackage{wrapfig}
\usepackage{todonotes}
\usepackage{blindtext}
\usepackage{caption}
\usepackage{subcaption}
\usepackage{siunitx}
\usepackage{pifont} 
\usepackage{tabularx}
\usepackage{xcolor}
\usepackage{xspace}
\usepackage{enumitem}


\newcommand{\ourmodel}{GraspGen\xspace}
\newcommand{\cmark}{\ding{51}}
\newcommand{\xmark}{\ding{55}}

\definecolor{myGreen}{RGB}{0, 185, 0}
\definecolor{myLightBlue}{RGB}{40, 214, 253}
\definecolor{myDarkBlue}{RGB}{65,94,242}

\newcommand{\graspgen}{GraspGen\xspace}
\makeatletter\renewcommand\paragraph{\@startsection{paragraph}{4}{\z@}
  {.5em \@plus1ex \@minus.2ex}{-.5em}{\normalfont\normalsize\bfseries}}\makeatother

\title{GraspGen: A Diffusion-based Framework for 6-DOF Grasping with On-Generator Training}

\author{
Adithyavairavan Murali \hspace{5px} Balakumar Sundaralingam \hspace{5px} Yu-Wei Chao \hspace{5px} Wentao Yuan$^*$ \hspace{5px} Jun Yamada$^*$ \hspace{20px} Mark Carlson \hspace{5px} Fabio Ramos \hspace{5px} Stan Birchfield \hspace{5px} Dieter Fox$^*$ \hspace{5px} Clemens Eppner$^*$
\\
\tiny{* Work done at NVIDIA} \\
\large{\textbf{\url{https://graspgen.github.io}}}
\\
}

\begin{abstract}
Grasping is a fundamental robot skill, yet despite significant research advancements, learning-based 6-DOF grasping approaches are still not turnkey and struggle to generalize across different embodiments and in-the-wild settings. We build upon the recent success on modeling the object-centric grasp generation process as an iterative diffusion process. 
Our proposed framework, \textit{GraspGen}, consists of a Diffusion-Transformer architecture that enhances grasp generation, paired with an efficient discriminator to score and filter  sampled grasps. We introduce a novel and performant on-generator training recipe for the discriminator. To scale GraspGen to both objects and grippers, we release a new simulated dataset consisting of over 53 million grasps. We demonstrate that GraspGen outperforms prior methods in simulations with singulated objects across different grippers, achieves state-of-the-art performance on the FetchBench grasping benchmark, and performs well on a real robot with noisy visual observations.
\end{abstract}

\begin{document}

\makeatletter
    \let\@oldmaketitle\@maketitle
    \renewcommand{\@maketitle}{\@oldmaketitle
    \begin{minipage}[l]{0.65\textwidth}
        \centering
        \includegraphics[width=\linewidth]{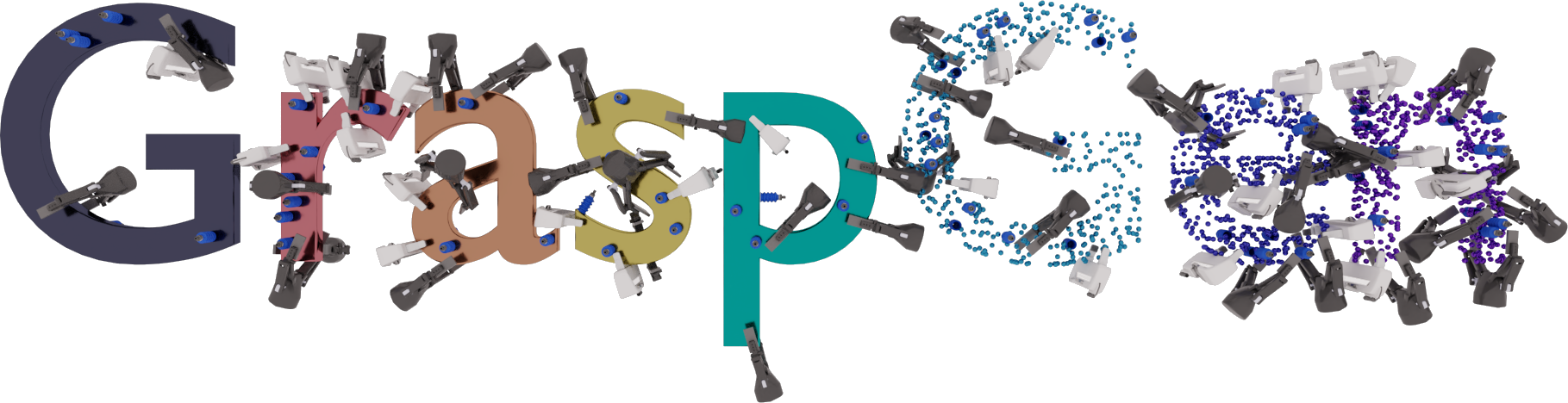}
    \end{minipage}
    \hfill
    \begin{minipage}[r]{0.3\textwidth}
        \centering
        \includegraphics[width=\linewidth]{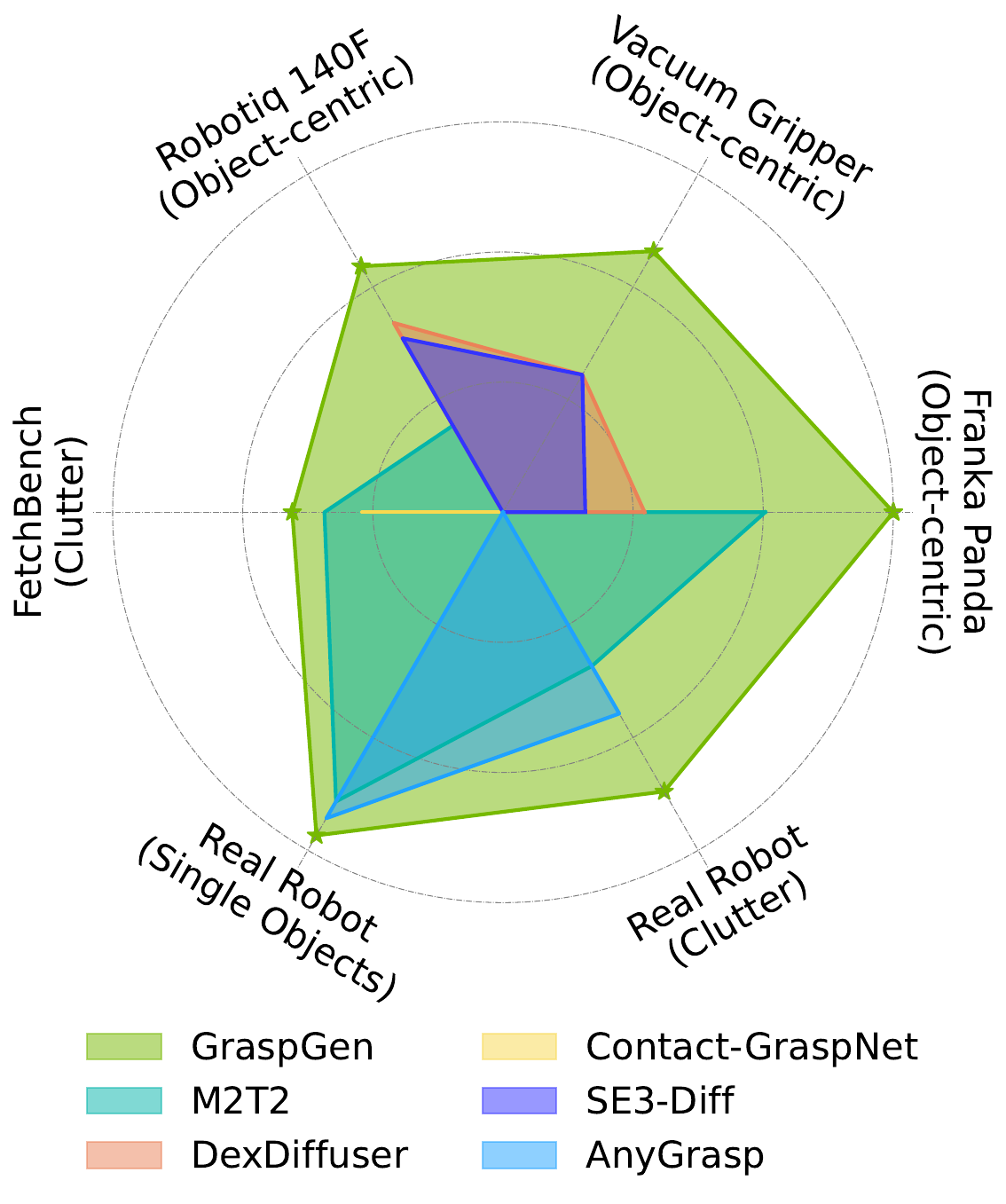}
    \label{fig:teaser}
    \end{minipage}
     \caption{We present \ourmodel, a framework for learning and datasets for 6-DOF grasping in multiple settings and embodiments.}
  \vspace{-25pt}
    }
\makeatother
\setcounter{figure}{1}
\maketitle

\abscontent
\section{Introduction}
\label{sec:introduction}
\begin{wrapfigure}{r}{0.4\textwidth}
    \centering
    \vspace{-15pt}
    \includegraphics[width=\linewidth]{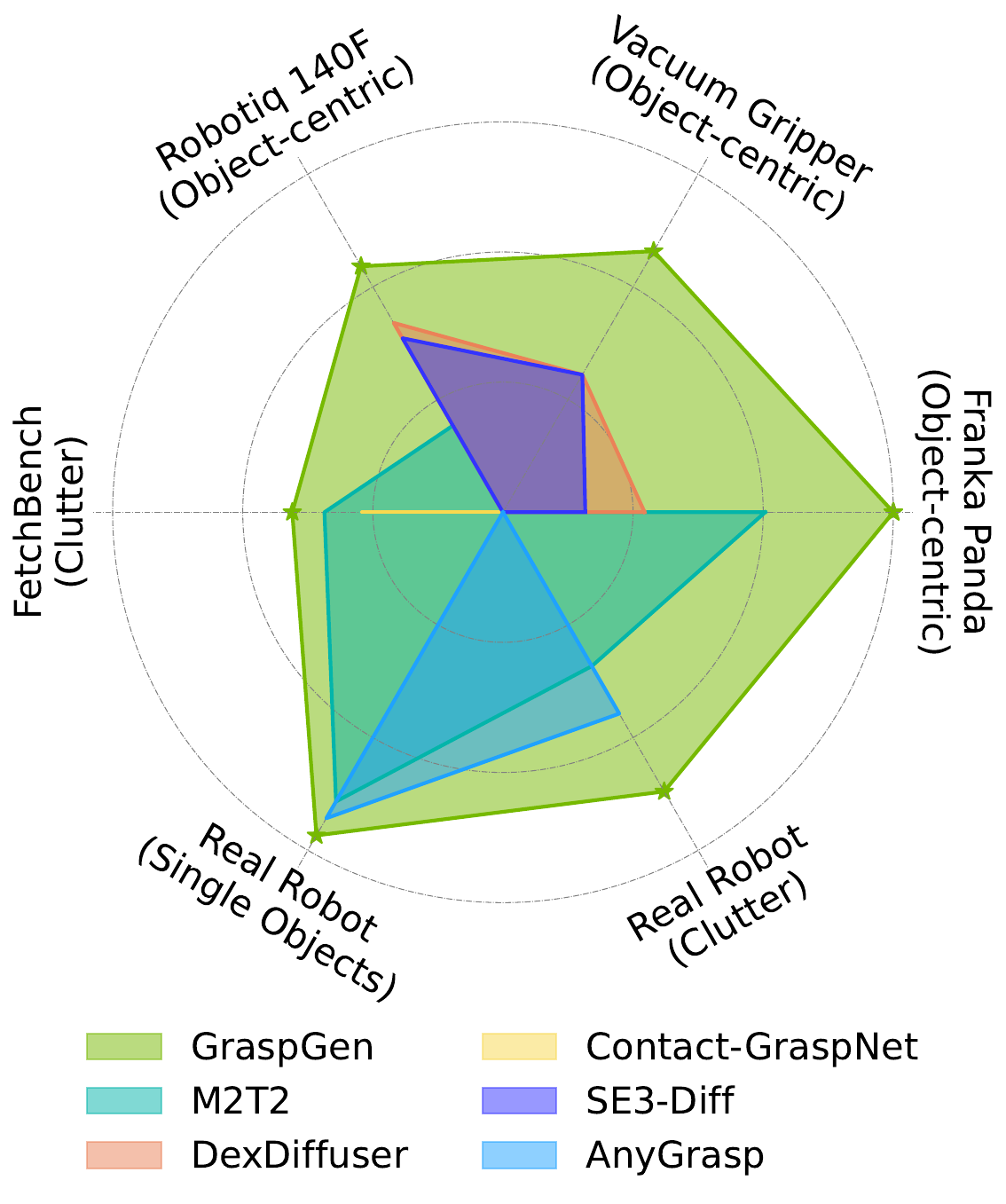}
    \vspace{-2em}
\end{wrapfigure}
Robot grasping has seen significant advances in recent years: including data generation \citep{eppner2021acronym}, generalization across embodiments~\citep{xu2021adagrasp}, integration with touch sensing~\citep{murali2018iser, calandra2018more}, operating in complex cluttered environments \citep{Murali2020CollisionNet}, language prompting \citep{murali2020taskgrasp,tang2023graspgpt}, and real-world RL algorithms \citep{qtopt2018}.

However, recent results show that critical gaps still exist in the development of a general-purpose grasping system.
In the FetchBench~\citep{han2024fetchbench} benchmark, state-of-the-art (SOTA) grasping systems achieve sub-20$\%$ accuracies.
Similarly, the OK-Robot effort~\citep{liu2024okrobot}, which introduced a knowledge-based system for mobile manipulation in-the-wild, reported a notable error rate of 8$\%$~(30 errors out of 375 trials) due to grasp model failures alone. 
These grasping models perform at approximately~60$\%$ accuracy in their evaluations.
The evaluation of Robo-ABC~\citep{ju2025robo}~(which uses a SOTA grasp method) as a baseline in RAM~\citep{kuang2024ram} shows sub-50$\%$ success rates.
This highlights the need for further advancements in grasping frameworks to ensure their reliability as subroutines in higher-level reasoning systems~\citep{liu2024okrobot,dalal2024manipgen, deshpande2025graspmolmo, huang2024rekep}.

Furthermore, grasping systems are not yet turnkey, and making them more flexible remains an open systems research challenge.
For instance, classical model-based approaches to grasp generation required precise object pose information~\citep{deng2020self} which does not generalize for the unknown object setting. Other methods necessitate multi-view scans for a single object~\citep{lum2024get}, making them impractical for cluttered environments. Contact point-based architectures~\citep{sundermeyer2021-contact-graspnet,yuan2023m2t2}, often struggle to generalize to different gripper morphologies, limiting their applicability to hardware beyond symmetric parallel-jaw grippers. We also demonstrate that they have comparatively worse scoring of predicted grasps.

Although some methods have been proposed to generate grasps in cluttered environments with multiple objects~\citep{sundermeyer2021-contact-graspnet, yuan2023m2t2, fang2023anygrasp}, they typically require simulating entire scenes or manual data collection in the real world. This approach is challenging to scale to larger scenes beyond tabletops and raises questions about how to synthetically generate cluttered scenes that accurately represent real-world distributions at test time.
These methods yet still rely on instance segmentation for target-driven grasping.
However, recent advances in instance segmentation using foundation models like SAM2~\citep{ravi2024sam2} mitigate the need for world-centric models. This shift allows us to revisit and emphasize object-centric models, simplifying grasp generation during both training and inference.

In this work, we propose a new framework \textit{\graspgen} that achieves superior grasping performance compared to prior approaches. Our model is based on a combination of a diffusion-based generator and an efficient discriminator. Our technical novelty is two-fold. First, we show that GraspGen is a flexible system for scaling grasp generation across diverse settings, including: \textit{embodiments} (compatibility with 3 distinct gripper types), \textit{observability} (robustness to partial vs. complete point clouds), \textit{complexity} (single-object vs. cluttered scenes), and \textit{sim vs. real}. Second, we propose a novel training recipe (Algorithm \ref{alg:graspgen_training_recipie}). A highlight of this recipe is that the grasp discriminator is supervised with our On-Generator Dataset. No prior work using 6-DOF grasp discriminators ~\citep{mousavian2019-6dofgraspnet, weng2024dexdiffuser, PointNetGPD2019} have shown this and we demonstrate that On-Generator Training substantially improves the performance over a standard grasp discriminator trained with only offline data. Compared to prior work, our discriminator is aware of the mistakes made by the diffusion model and assigns a lower corresponding score for potentially false positive grasps. We show how different design choices, from our training recipe to architectural changes, improve on earlier works. Apart from grasp accuracy \graspgen enhances inference time and memory usage. We additionally provide a dataset consisting of 53 Million grasps, to support future research on these topics within the community.

\section{Related Work}

\textbf{6-DOF Grasping.} Planning robot grasps is usually formulated as a 6-DOF grasp pose detection problem~\citep{newbury2022review}, with components for both \textit{grasp sampling} (GS) and \textit{grasp analysis} (GA). Recently, generative models such as autoregressive models \citep{tobin2018grasp}, Variational Autoencoder (VAE) \citep{mousavian2019-6dofgraspnet} and diffusion models \citep{wu2023learning-dex-human-affordance, urain2022se3dif, lum2024get} have been proposed for GS. GA is typically done with a discriminator model to score and rank the sampled grasps \citep{weng2024dexdiffuser, mousavian2019-6dofgraspnet, song2024implicitgraspdiffusion, Murali2020CollisionNet}. Some methods have a single model for both GA and GS for efficient inference - \citep{sundermeyer2021-contact-graspnet} proposed a contact point grasp representation and \citep{yuan2023m2t2} extended this with a transformer for grasping as well as placing. Other works have investigated the choice of input modality, be it a 3D point cloud~\citep{sundermeyer2021-contact-graspnet,Murali2020CollisionNet, mousavian2019-6dofgraspnet}, an implicit representation \citep{lum2024get} or voxelization of the scene~\citep{breyer2020volumetric,jiang2021synergies}. Our framework requires an object-centric point cloud input. 

\textbf{Applications of 6-DOF Grasping.} Understanding the applications of 6-DOF Grasping is crucial to designing the right modular framework for this problem. Popular applications that use 6-DOF grasp networks as a submodule include target-driven grasping in clutter \citep{chen2024regionawaregraspframeworknormalized, xie2024rethinking6dofgraspdetection, sundermeyer2021-contact-graspnet, Murali2020CollisionNet}, dynamic grasping \citep{fang2023anygrasp} and language-guided semantic manipulation ~\citep{fang2020learning, murali2020taskgrasp, tang2024foundationgraspgeneralizabletaskorientedgrasping} (e.g. grasping a mug by its handle for pouring). Given the maturation of instance segmentation models such as SAM2\citep{ravi2024sam2}, we  directly reason with object-centric point cloud input, circumventing the need for scene modeling during training which is cumbersome from a data generation perspective. Since downstream knowledge systems \citep{dalal2024manipgen} may require these networks to work with either single-view camera observations (in constrained environments) or multi-view setups, we design our framework to handle both scenarios.

\textbf{Diffusion Models in Robot Manipulation.} Diffusion models \citep{ho2020denoising, song2019} are a powerful class of generative models. Recently the robotics community has applied them to a host of problems involving high-dimensional, multi-modal and continuous distributions: visuomotor policy learning \citep{chi2024diffusionpolicyvisuomotorpolicy, ke20243ddiffuseractorpolicy}, grasping \citep{urain2022se3dif, weng2024dexdiffuser, carvalho2024graspdiffusionnetworklearning, freiberg2024diffusionmultiembodimentgrasping, zhang2024diffgraspwholebodygraspingsynthesis, lum2024get}, motion planning \citep{huangdiffusionseeder}, rearrangement \citep{structdiffusion2023}, scene generation \citep{chen2024urdformer}, amongst many others. The closest paper to our work is \citep{urain2022se3dif} which proposed the problem formulation of 6-DOF antipodal grasping as a diffusion process for known objects (without point cloud input) and \citep{weng2024dexdiffuser} which extended the former to dexterous grasping from point cloud observations and added a discriminator for grasp analysis. In our framework, we provide a new large-scale multi-gripper dataset and improve upon both GA and GS.

\section{\ourmodel}
The objective of grasp generation is to synthesize a large spatially-diverse set of successful grasp poses. We need the grasps to be diverse for performant execution in clutter, where many otherwise successful grasps are unreachable or in collision and hence are filtered out at inference time by the motion planner. In practice, the required output of the grasp generation is a set of top-$K$ grasps for the object. Generated grasps need to be scored and ranked to return the best performing grasps. This is done with grasp evaluation in Sec \ref{sec:grasp_evaluation}.

\subsection{Grasp Generation with Diffusion}
\label{sec:diffusion}
We formulate the problem of 6-DOF grasp generation as a diffusion model in SE(3)~\citep{urain2022se3dif}. For a specific object, the grasp distribution is continuous and highly multimodal, making it a suitable problem for generative modeling.
At a high level, diffusion models entail adding noise sequentially to the training data.
This process is reversed during inference time, where the data is generated from noise.
Urain et al.~\citep{urain2022se3dif} proposed to learn an energy-based model (EBM) with score-matching Langevin dynamics~(SMLD)~\citep{song2019}.
Inference sampling requires computing the logarithmic probability gradient of the EBM network, which is computationally slow.
Instead, we formulate the problem as a Denoising Diffusion Probabilistic Model~(DDPM)~\citep{ho2020denoising}, which models a distribution using an iterative denoising process. DDPM is empirically faster to compute and simpler to implement.
Recent work has demonstrated the equivalence between the paradigms~\citep{song2021b}. The space on which we perform the diffusion is in the SE(3) Lie group.
Unfortunately, the rotation space is not a Euclidean, but DDPMs are proposed to model data coming from a Euclidean space in $\mathbb{R}^{n}$.
Analogous to~\citep{urain2022se3dif} we factorize SE(3) into SO(3) $\times $ $\mathbb{R}^{3}$, where $\mathbb{R}^{3}$ and SO(3) Lie algebra spaces are Euclidean.
We use a conditional diffusion model since the noise prediction network is conditioned on a point cloud encoding the object shape.

\begin{figure}
  \begin{center}
    \includegraphics[width = 0.89\linewidth]{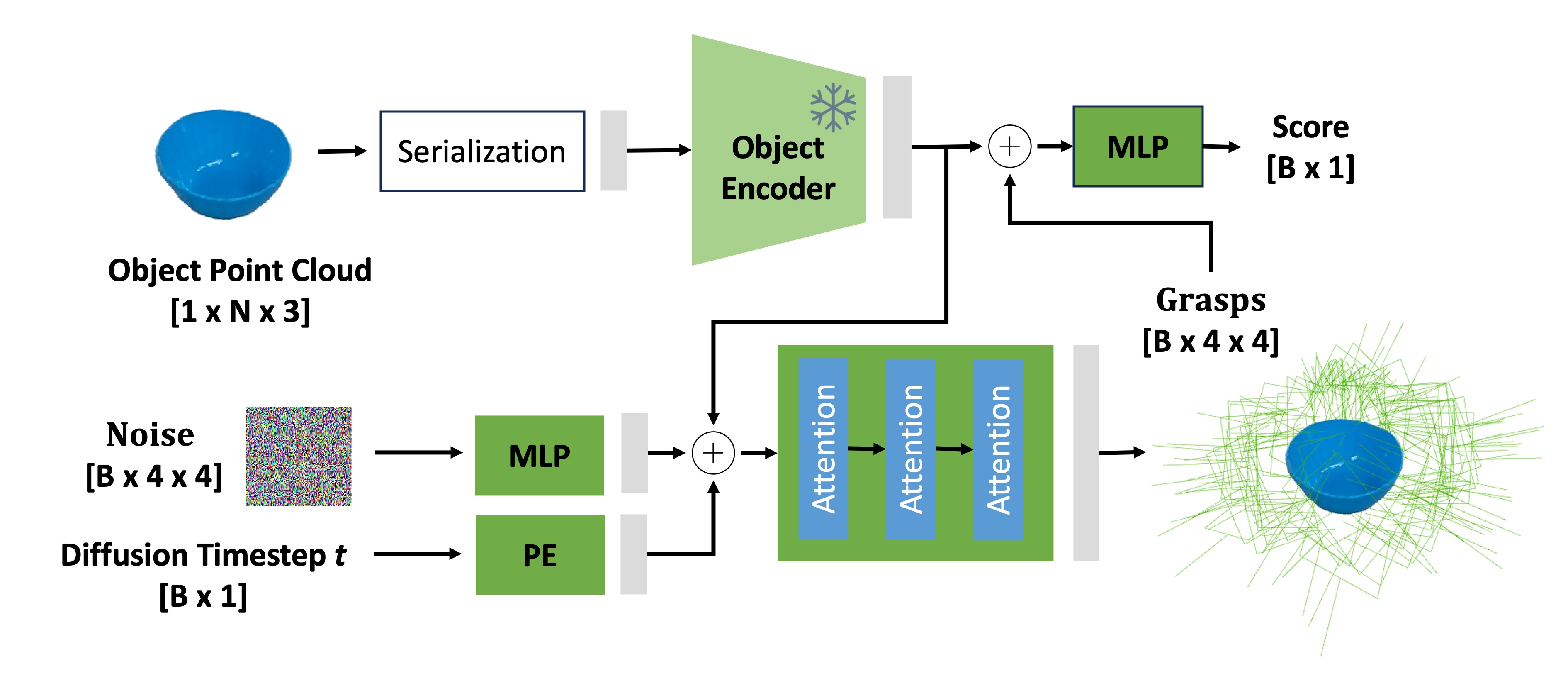}
  \end{center}
  \caption{Architecture for the diffusion noise prediction network.}
  \vspace{-5mm}
\label{fig:network}
\end{figure}

\textbf{Translation Normalization.}
Neural networks perform best when the data is normalized and input and outputs are properly scaled.
For SO(3), the space is bounded between $[-\pi, \pi]$.
However, translation is unbounded and heavily dependent on the scale of the object point cloud.
While the object point clouds can be rescaled to be within a bound, the bounds of grasps in SE(3) vary based on each object's shape and pose.
We normalize the grasp translations by the multiplier $\kappa$. Instead of setting this value arbitrarily or with a grid search, we compute this from the dataset statistics as $\kappa= \frac{1}{\frac{1}{N} \sum_{i=0}^{N} \left( max(t_{i}) - min(t_{i}) \right) }$ where $t_{i}$ is the translation component in $\mathbb{R}^{3}$ of all the positive grasps poses $\mathcal{G}_{i}^{+}$ for object $i$. 

\textbf{Object Tokenization.}
PointNet++~\citep{pointnet} is a popular choice as an object encoder for several 6-DOF grasping papers~\citep{mousavian2019-6dofgraspnet, urain2022se3dif, murali2020taskgrasp, PointNetGPD2019}.
While point cloud transformer architectures~\citep{wu2022ptv2} are making steady progress, to the best of our knowledge, no generative grasping paper has used them to encode objects.
We use the recently proposed PointTransformerV3~(PTv3)~\citep{wu2024ptv3} as a backbone.
PTv3 uses serialization to convert unstructured point clouds to a structured format, before applying a transformer on this serialized output.
This sidesteps the process of nearest neighbor ball query search for hierarchical feature processing, a common bottleneck in prior point cloud processing frameworks.

\textbf{Diffusion Network.} The diffusion noise prediction network, as used during inference time, is shown in Fig. \ref{fig:network}. Both the point clouds as well as the grasps are transformed to the point cloud mean center before passing through the noise prediction network. During inference, we first sample a noise vector (with batch size $B$) and iteratively execute the diffusion reverse process to generate the grasps.
Through hyper-parameter search we found that $T=10$ denoising steps are sufficient for our setting.
While diffusion models that generate images typically run for over 100s of steps, we hypothesize that diffusion on grasps should be less complex, since the grasp dimensionality~(3 for translation + 3 for rotation) is significantly lower than the dimensionality of pixels and videos~($>50K$ for a $224 \times 224$ image).
The training loss is a denoising loss on the position and orientation difference between the predicted vs. actual noise values: $L = \left \lVert \epsilon - \phi(t, \tilde{g}, \mathcal{X}) \right \rVert_2^{2}$.
Here $\phi$ is the noise prediction network and $\mathcal{X}$ is the object point cloud. During training, we sample a random diffusion time step~$t \in [0, T]$ and add random noise~$\tilde{g} = g + \epsilon$ to the ground truth grasp~$g \in \mathcal{G}^{+}$. The diffusion timesteps and grasp poses are processed with position encoding and a multilayer perceptron respectively. We empirically found that running two separate denoising processes with their own dedicated scheduler yielded better performance than running a single DDPM for the translation and rotation components.
Note that the grasp scoring with the discriminator, as explained in the next section, is trained separately and is not used during diffusion model training.

\subsection{Grasp Evaluation with On-Generator Training}
\label{sec:grasp_evaluation}
A generative model trained solely on successful grasp data is prone to generating false positives due to model fitting errors.
In practice, a mechanism is needed to score, rank and filter each grasp before executing it on the robot.
To address this problem, earlier works~\citep{mousavian2019-6dofgraspnet, lum2024get, weng2024dexdiffuser} used a separately learned discriminator. We propose two key improvements.

\textbf{On-Generator Training.} Sim-to-real grasp models are typically trained with offline datasets of successful $\mathcal{G}^{+}$ and unsuccessful grasps $\mathcal{G}^{-}$. However, we show that the distribution of grasps from the generative model~$\hat{\mathcal{G}}$ is different from this offline dataset.
We believe this is due to the nature of the grasp sampling algorithm during training. For instance, the unsuccessful grasps may never collide with the object (which is the case for ACRONYM~\citep{eppner2021acronym}). However, some grasps generated by the diffusion model are slightly in collision with the object potentially due to model fitting errors.
Furthermore, some generated grasps are occasionally outliers and are far away from the object. These correspond to noise samples with low likelihoods. We hypothesize that such failure modes can all be removed with training a discriminator with On-Generator training, as described in Algorithm~\ref{alg:graspgen_training_recipie}.
We first run inference on the training set with the diffusion model.
This dataset corresponds to about 7K objects, each with 2K grasp samples per object. We then annotate this dataset by simulating the grasps using the same workflow used to generate the initial offline dataset.
This On-Generator dataset approximately corresponds to the original size of the initial offline dataset. 

\begin{wrapfigure}{R}{0.55\textwidth}
  \vspace{-23pt}%
  \begin{minipage}{0.55\textwidth}
    \begin{algorithm}[H]
    \small 
    \textbf{Given:} Object dataset $\mathcal{O}$, Grasp dataset $\{ \mathcal{G}^{+}$, $\mathcal{G}^{-} \}$ \\
    \textbf{Compute Translation Normalization:} $\kappa \leftarrow trans\_norm(\mathcal{G}^{+})$ \\
    \textbf{Train Generator:} $\pi^{gen} \leftarrow train\_DDPM(\mathcal{O}, \mathcal{G}^{+}, \kappa)$ \\
    \textbf{Sample On-Generator dataset:} $\hat{\mathcal{G}} \sim$
    $\pi^{gen}$($\mathcal{O}$) \\
    \textbf{Annotate On-Generator dataset:} $ \{ \hat{\mathcal{G}}^{+}, \hat{\mathcal{G}}^{-} \} \leftarrow$ simulate($\mathcal{O}$, $\hat{\mathcal{G}} $) \\
    \textbf{Train Discriminator:} $\pi^{dis} \leftarrow train\_classifier(\{\hat{\mathcal{G}}^{+}, \hat{\mathcal{G}}^{-}\}, \pi^{gen},\kappa)$ \\
    \textbf{Return:} ($\pi^{gen}, \pi^{dis}, \kappa$)
    \caption{GraspGen Training Recipe}
    \label{alg:graspgen_training_recipie}
    \end{algorithm}
  \end{minipage}
\end{wrapfigure}

\textbf{Efficient Evaluation.} Prior discriminator architectures~\citep{mousavian2019-6dofgraspnet, lum2024get} had their own object encoder separately trained from scratch. We propose a simpler architecture, which reuses the object encoder from the generation stage for the subsequent grasp discrimination step. As shown in Fig.~\ref{fig:network}, an MLP takes in this object embedding and a corresponding grasp pose and predicts a sigmoid score of grasp success. Another important design decision is about efficiently combining the embedding for the object shape and grasp pose. In prior work~\citep{mousavian2019-6dofgraspnet}, the grasp pose in SO(3) was converted to a point cloud (a handful of canonical points on the gripper were predefined and transformed with the grasp pose), concatenated with the object and passed into a PointNet with an additional input of a segmented point cloud.
Instead, we simply concatenate the object embedding with a SO(3) $\times $ $\mathbb{R}^{3}$ representation of the grasp pose. The discriminator is trained separately from the diffusion-based generator. Only the final MLP layer is trained from scratch with a binary cross entropy loss. The object encoder from the generator is fronzen and re-used for the discriminator.

\subsection{Dataset}
\label{sec:dataset}

Our dataset includes 6D gripper transformations and corresponding binary success labels for a repertoire of object meshes.
The label generation process follows the pipeline used in ACRONYM~\citep{eppner2021acronym}.
While ACRONYM is based on ShapeNetSem~\citep{savva2015semantically}, we use the more permissive, larger, and more diverse Objaverse dataset of 3D objects~\citep{deitke2023objaverse}.
Specifically, we select a subset of meshes from Objaverse that overlap with the 1,156 categories in the LVIS dataset~\citep{gupta2019lvis} and are licensed under CC BY~\citep{ccby}, totaling 36,366 meshes.
To compare with models trained on ACRONYM, we further select a random subset of 8,515~object meshes to match the size to the ACRONYM dataset.
For each object, 2K grasp transformations are uniformly sampled around the mesh. The label of a grasp is determined by simulating a shaking motion with the object in hand in the Isaac simulator~\citep{issacgym2021}. A grasp is considered successful if a stable contact configuration is present after the shaking motion finishes.
We construct datasets accordingly for the Franka Panda gripper and Robotiq-2f-140.
We generate a similarly structured dataset for a vacuum gripper (30mm suction cup) where success is labeled using an analytical model~\citep{mahler2018dexnet30computingrobust}.
Each gripper comprises $\approx$ 17M grasps.
\section{Experimental Evaluation}

\subsection{Simulation Results} 

\textbf{Baseline Methods.} We compare \ourmodel with multiple recent methods: Contact-point architectures M2T2~\citep{yuan2023m2t2} and Contact-GraspNet~\citep{sundermeyer2021-contact-graspnet}, diffusion architectures DexDiffuser~\citep{weng2024dexdiffuser} and SE3-Diffusion Fields~\citep{urain2022se3dif} as well as AnyGrasp~\citep{fang2023anygrasp}. We reimplemented SE3-Diffusion Fields with two key differences: (1) a PointNet++ backbone to process the point cloud input of unknown objects and (2) model trained with DDPM~\citep{ho2020denoising} instead of SMLD~\citep{song2019}. Since the model does not score each generated grasp, we use the approximate log-likelihood instead. 
DexDiffuser~\citep{weng2024dexdiffuser} was originally proposed for dexterous grasping, where grasps are parameterized by a pose and gripper joint configuration. Since we focus on pinch and suction grasping, we reuse the architecture with only the pose input. We skip comparing to a Variational Auto-encoder (VAE) baseline since it was already demonstrated to have worse performance than the baselines we compared to ~\citep{urain2022se3dif, sundermeyer2021-contact-graspnet} in their corresponding papers. We directly compare to these more recent baselines instead of repeating the VAE baseline.

\begin{wrapfigure}{r}{0.45\textwidth}
    \centering
    \vspace{-10pt}%
    \includegraphics[width=\linewidth]{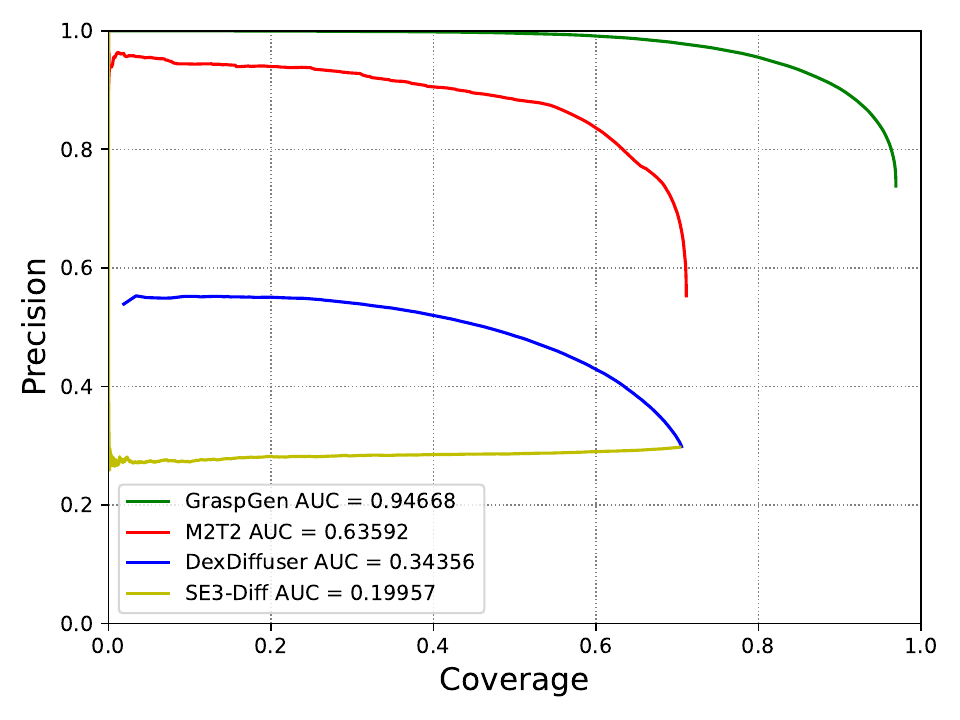}
    \caption{Object-centric evaluation on Franka-ACRONYM ~\citep{eppner2021acronym}}
    \label{fig:baseline_comparison}
\end{wrapfigure}
\textbf{Full Point Cloud of Single Objects.}
We begin by evaluating grasp generation using a complete point cloud sampled from the object's mesh, without any self-occlusion. For each method, we evaluate on a test set of 815 objects with 2K grasps/object, resulting in a total of 1.6 million grasp executions.
We trained all models using the same training set and splits. For M2T2, we just use a single transformer token, since there is only one object in the scene. The evaluation pipeline follows the setting of data collection in Sec.~\ref{sec:dataset}, where the grasp poses are attempted with a isolated free-floating gripper. We present results on the widely used ACRONYM dataset \citep{eppner2021acronym} for the Franka gripper. Extended results on \ourmodel datasets are presented in the Appendix. In this section, we skip comparing with Contact-GraspNet~\citep{sundermeyer2021-contact-graspnet} since this was already shown to be worse in the M2T2 paper~\citep{yuan2023m2t2} and we further compare it in Sec.~\ref{sec:fetchbench}. Due to license restrictions limiting model deployment to registered machines, we were unable to compare to AnyGrasp ~\citep{fang2023anygrasp} in these simulation experiments on the compute cluster but evaluated it in the real world on a registered desktop.

The results are summarized in the Precision-Coverage curve in Fig.~\ref{fig:baseline_comparison}. \textit{Precision} represents the grasp success rate in simulation, while \textit{Coverage} is a measure of spatial diversity of the grasps and is the percentage of the ground truth positive grasp set, $\mathcal{G}^{+}$, matched by the predicted grasps. The matching is done using nearest neighbour assignment (distance of 1$cm$) used in prior work ~\citep{yuan2023m2t2, sundermeyer2021-contact-graspnet, mousavian2019-6dofgraspnet}. \ourmodel outperforms baselines by over \SI{48}{\percent} in terms of Area Under Curve (AUC). All methods with discriminative reasoning (\ourmodel, DexDiffuser, M2T2) outperformed SE3-Diff, a purely generative method and scored based on the approximate loglihood - cementing the importance of a discriminator in grasp generation. This result also demonstrates that the discriminator quality is crucial. \ourmodel with its On-Generator training is able to better score the grasps compared to the discriminator of DexDiffuser \citep{weng2024dexdiffuser}, as it specifically trains for the distribution of grasps from the diffusion model. While the discriminator in \ourmodel is trained on both positive and negative grasps, M2T2 is trained exclusively on positive grasps and only distinguishes between good and bad contact points, resulting in a worse performance.

\begin{figure*}[t]
 \centering
 \includegraphics[width=\linewidth]{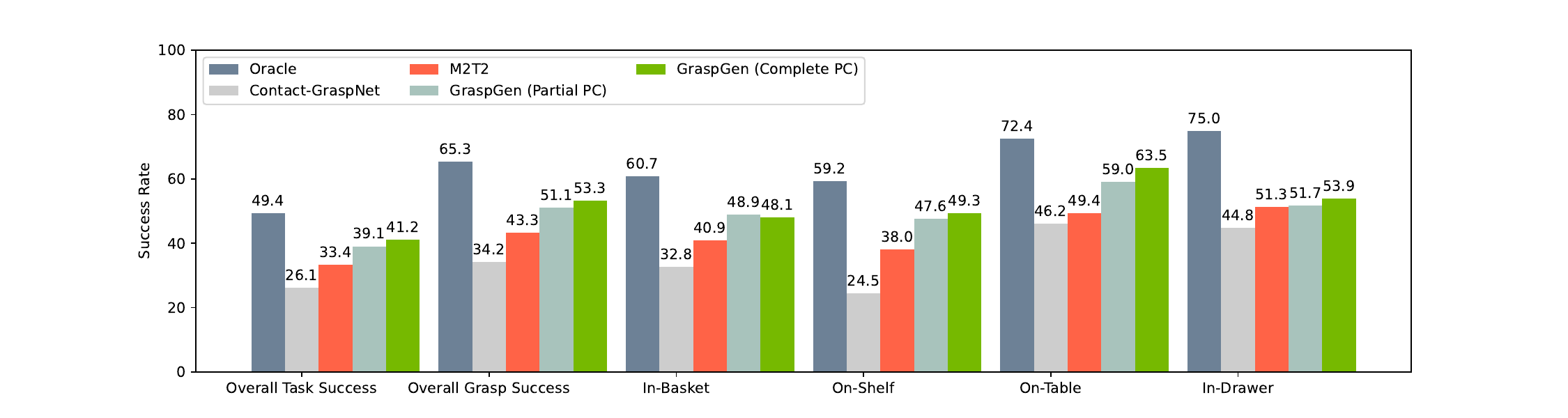}
\vspace{-2em}
 \caption{Large-scale evaluation on FetchBench \citep{han2024fetchbench}. \ourmodel surpasses all previous methods.}
  \vspace{-5mm}
 \label{fig:fetchbench}
\end{figure*}

\textbf{Task-level Evaluation in Clutter.}
\label{sec:fetchbench}
We evaluate \ourmodel's ability to handle complex grasping in clutter using FetchBench~\citep{han2024fetchbench}, a simulation-based grasping benchmark with diverse procedural scenes.
FetchBench simulates all stages of grasping, from perception, grasp pose detection, collision world modeling, to motion planning.
The experiments are conducted with a Franka Panda robot in 100 scenes~(see Fig.~\ref{fig:fetchbench}) with 60~tasks per scene for a total of 6k~grasp executions.
To focus solely on grasp pose detection and eliminate confounding factors, we use the ground truth collision mesh of the scene for motion planning with cuRobo~\citep{curobo_report23}.
We first report an \textit{oracle} planner with ground truth grasps from the dataset~\citep{eppner2021acronym}, to demonstrate the best grasp performance possible without any sensing or model performance issues.
We report two success metrics: 1) task success rate, which measures successful completion from grasping to placing the object, and 2) grasp success rate, which tracks successful grasps only.
The latter is always higher, as some grasps succeed but may slip or collide during retraction.
Interestingly, the \textit{oracle} planner achieves only \SI{65}{\percent} grasp success and \SI{49.2}{\percent} task success. This low performance stems from several factors: 1) many scenarios allow grasping but lack a collision-free retraction path, 2) some problems exceed the capabilities of existing motion planners~\citep{curobo_report23}, and 3) objects are in complex poses where stable grasps exist but are inaccessible. Addressing these challenges requires more advanced reasoning policies beyond the scope of this paper. Nonetheless, \ourmodel achieves SOTA results surpassing Contact-GraspNet~\citep{sundermeyer2021-contact-graspnet} and M2T2 ~\citep{yuan2023m2t2} by ~\SI{16.9}{\percent} and ~\SI{7.8}{\percent} respectively.

\begin{wrapfigure}{r}{0.6\textwidth}
    \centering
    \vspace{-10pt}%
    \includegraphics[width=0.48\linewidth]{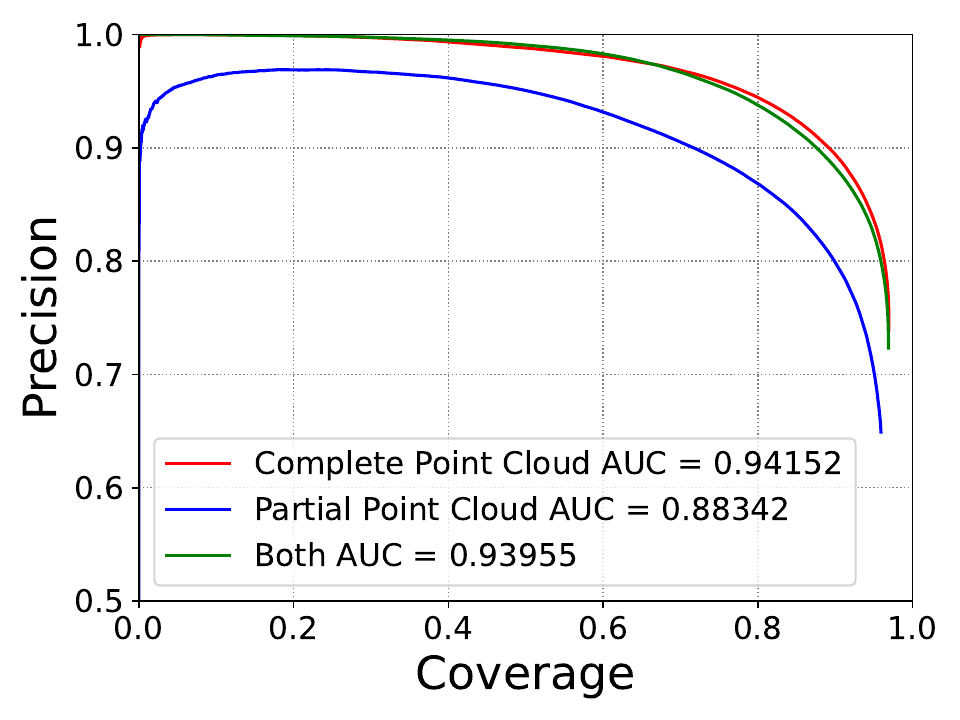}%
    \hfill%
    \includegraphics[width=0.48\linewidth]{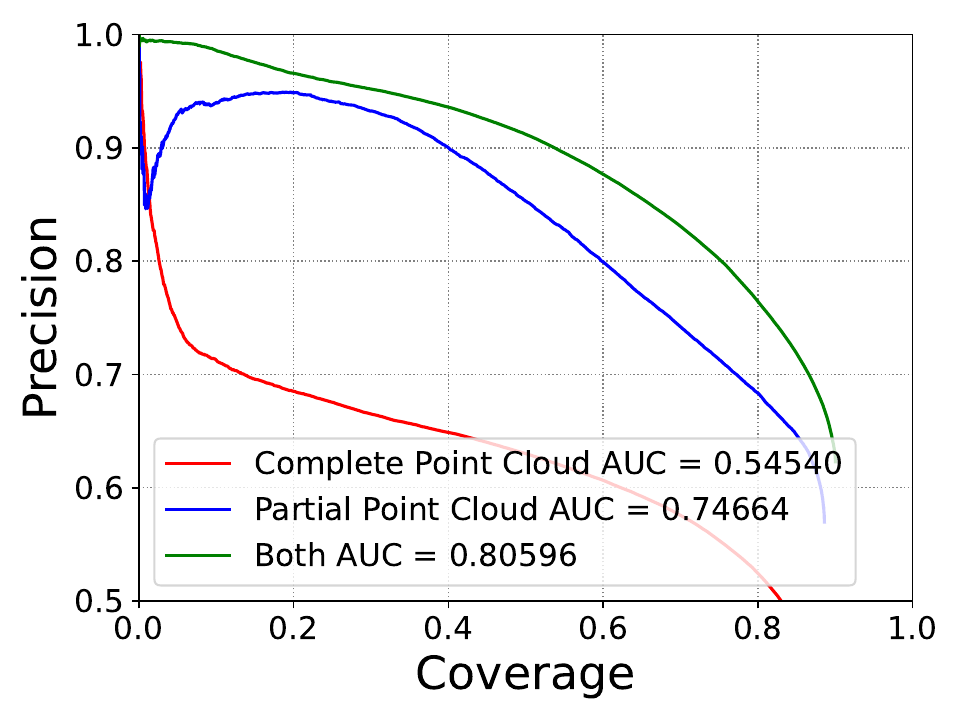}%
    \caption{Evaluating on complete \textit{(left)} vs. single-view point clouds \textit{(right)}}
    \label{fig:ablation_pc}
\end{wrapfigure}
\textbf{Sensitivity to Occlusions.}
Prior work has proposed 6-DOF grasp generation for either single-view partial point clouds with strong self-occlusion \citep{mousavian2019-6dofgraspnet, sundermeyer2021-contact-graspnet, yuan2023m2t2} or complete point clouds, obtained by fusing multiple camera views \citep{murali2020taskgrasp, urain2022se3dif, lum2024get}. As shown in Fig.~\ref{fig:ablation_pc}, \ourmodel trained on partial point clouds performs poorly on complete point clouds and vice versa. By training on a mix of both (50-50 split), \ourmodel generalizes across both settings, improving flexibility for downstream applications.

\subsection{Analysis of On-Generator Training}
\label{sec:ongenerator_analysis}
\begin{wrapfigure}{r}{0.6\textwidth}
    \centering
    \vspace{-20pt}%
    \includegraphics[width=0.48\linewidth]{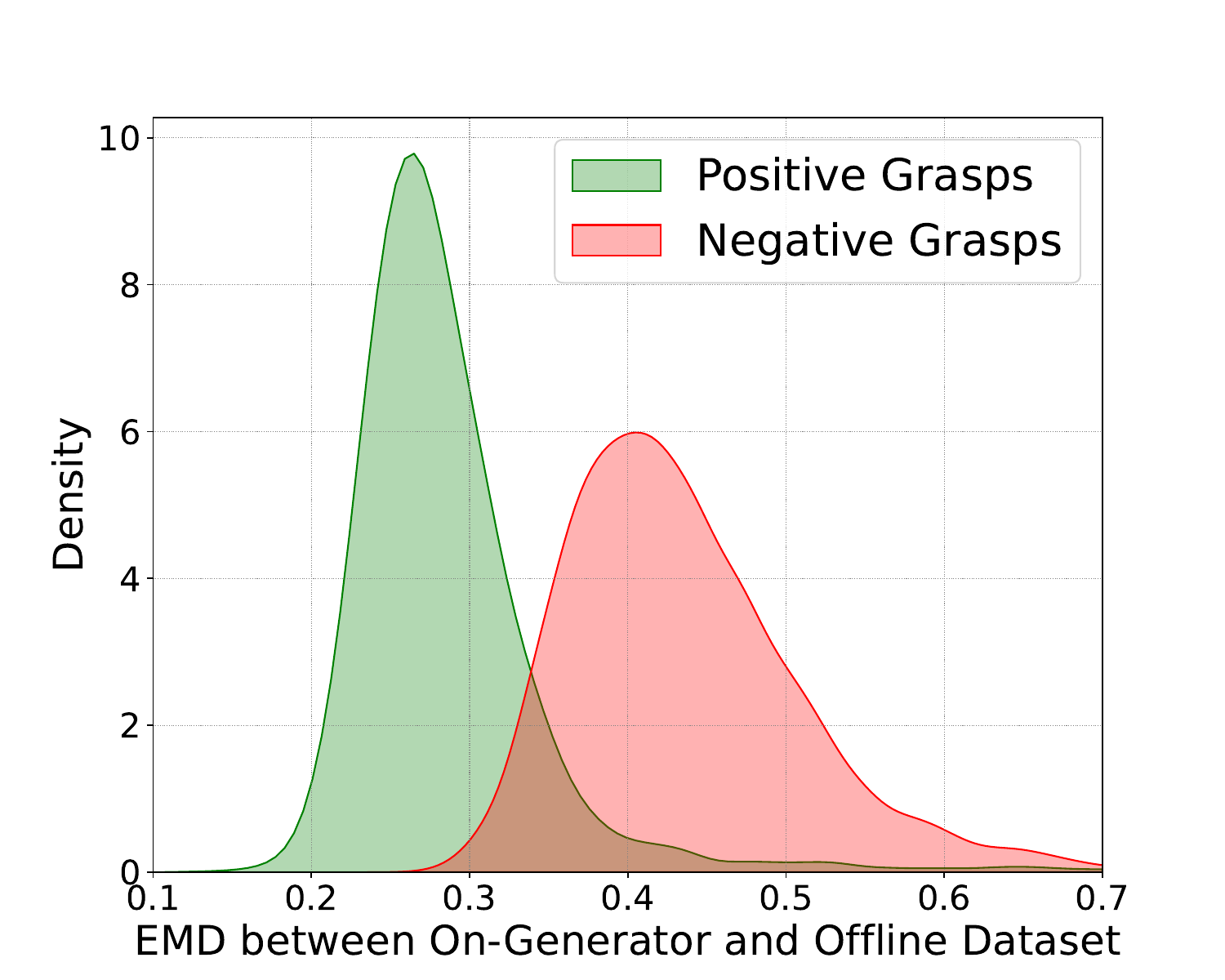}%
    \hfill%
    \includegraphics[width=0.48\linewidth]{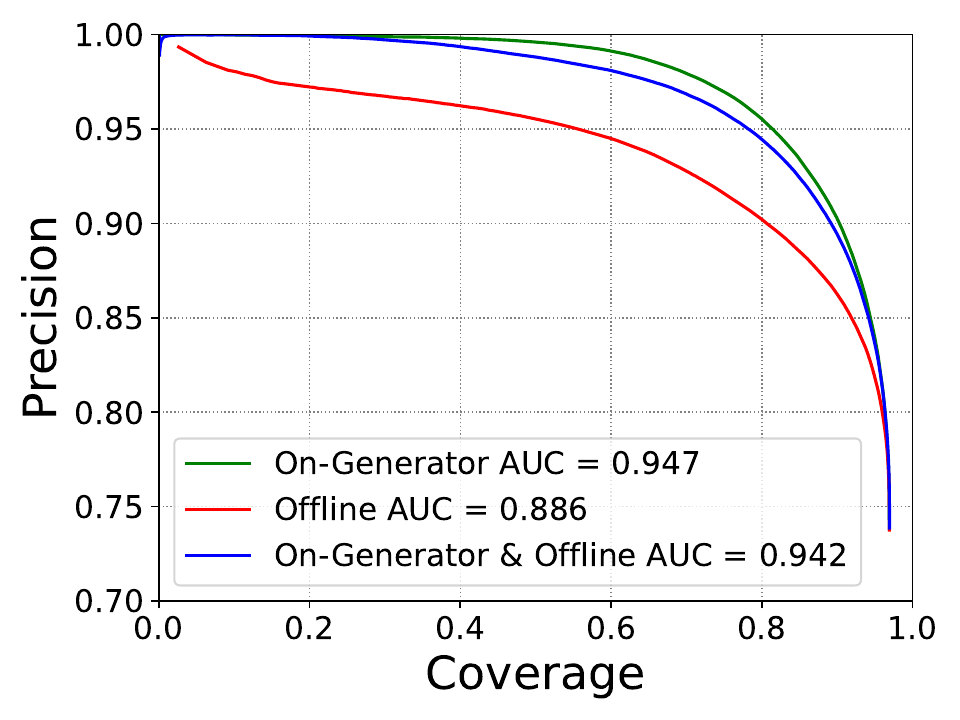}%
    \caption{Distribution Shift in the On-Generator vs. Offline Datasets \textit{(left)} and Ablation on Trained Models \textit{(right)}}
    \label{fig:dataset_emd_and_ongenerator_ablation}
\end{wrapfigure}
On-Generator training of the discriminator is essential for enabling the model to recognize its own failure modes and filter out erroneous predictions. Before we show results with On-Generator training, we first demonstrate that there is a measurable distribution shift between the offline data distribution vs. diffusion-model generated sample distribution. We use the Earth Mover's Distance (EMD) (detailed in the Appendix Sec.~\ref{supp:emd}) and plot this separately for positive and negative grasps for the entire training set of ($\sim$7K) objects in Fig \ref{fig:dataset_emd_and_ongenerator_ablation}(left). There is a substantial non-zero EMD between the offline and on-generator datasets. It is especially more pronounced for the negative grasps, since the spatial manifold of unsuccessful grasps (e.g. grasps far away or colliding an object are still considered negative examples) is larger than that of successful grasps (i.e. grasps need to be close to the object surface). This distribution shift justifies our need to train our discriminator to specifically filter out unsuccessful grasps.

As shown in Figure \ref{fig:dataset_emd_and_ongenerator_ablation}, the model trained exclusively on On-Generator data achieved the highest performance. The model trained with the offline dataset performed the worst (\SI{6.5} {\percent} lower AUC). We hypothesize that the On-Generator training captures the false positives of the diffusion model better than the offline dataset. For example, the generator may produce grasps that are slightly in collision with the object's collision mesh -- cases absent from the offline dataset, which contains only collision-free grasps.
Additionally, model fitting errors can introduce outliers with small translation or rotation errors, leading to unstable grasps.

\subsection{Ablation Studies}
\textbf{Analysis of the Discriminator.}
In comparison to SOTA architectures~\citep{mousavian2019-6dofgraspnet}, our discriminator is more accurate~(6.7$\%$ higher AUC) and uses 21$\times$ less memory. See Appendix~\ref{appendix:discriminator} for details.

\begin{wrapfigure}{r}{0.5\textwidth}
    \centering
    \includegraphics[width=\linewidth]{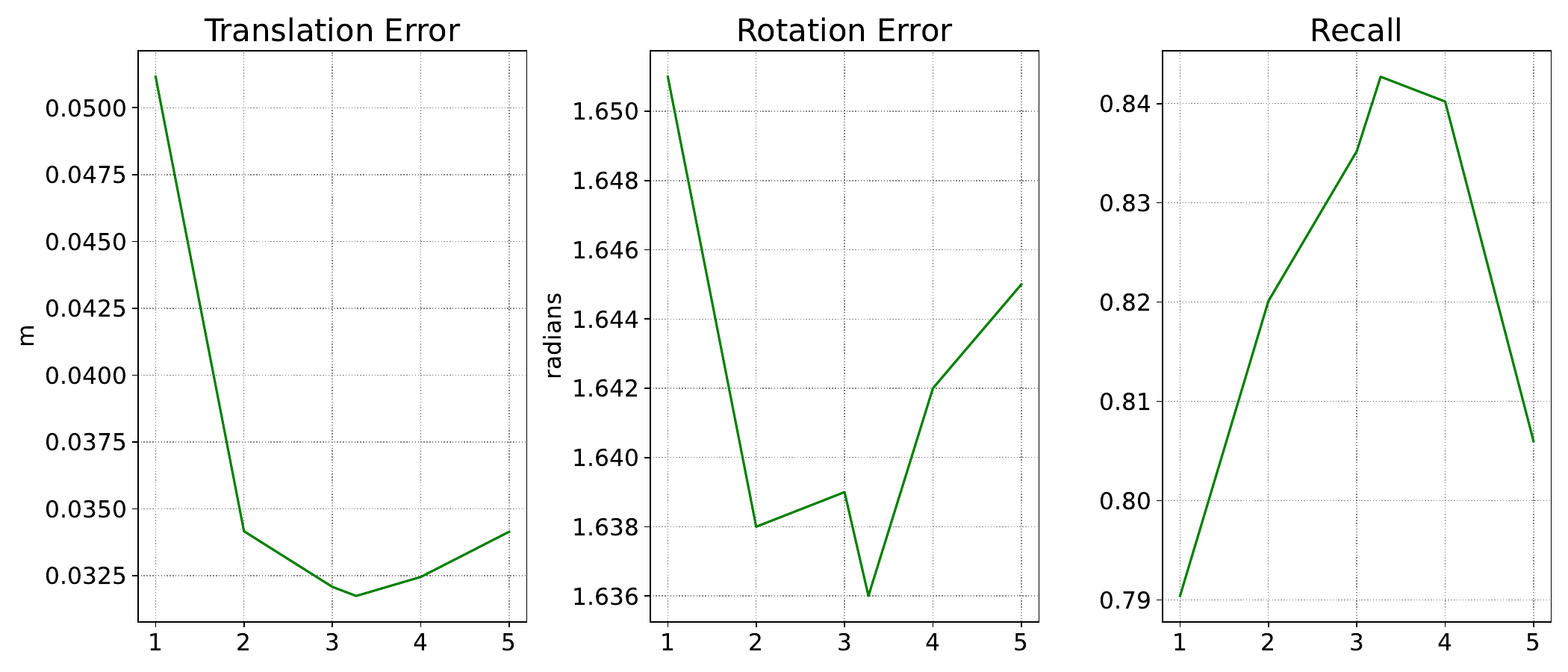}
    \caption{Ablation on Translation Normalization on the Franka-ACRONYM ~\citep{eppner2021acronym} dataset. $\kappa$ is plotted on the x-axis}
    \label{fig:ablation_gen_scale}
\end{wrapfigure}
\textbf{Translation Normalization.} We found a convex relationship between the performance of the diffusion model and the normalization multiplier. Fig.~\ref{fig:ablation_gen_scale} summarizes the key results, where the goal is to minimize the translation/rotation error while maximizing the coverage (recall). 
While an optimal scale factor can be found via hyperparameter grid search, we empirically observe that computing the normalization multiplier using the equation detailed in Sec \ref{sec:diffusion}) results in a local minima and provides a reliable alternative. $\kappa$=3.27 for Franka gripper.

\textbf{Rotation Representation.}
All tested representations -- 6D rotation representation, Euler angles, Lie Algebra --  performed comparably. For experimental details see Appendix~\ref{appendix:rotation}.

\textbf{Pointcloud Encoder.}
We demonstrate substantial gains using the SOTA transformer backbone PointTransformerV3~(PTv3)~\citep{wu2024ptv3} over PointNet++~\citep{pointnet}. PTv3 reduces translation error by~\SI{5.3}{mm} and increases recall by~\SI{4}{\percent}. See Appendix~\ref{appendix:pointcloudencoder} for more details.

\subsection{Performance on Multiple Grippers}
While the main paper presents comparisons for the Franka Panda gripper, results for the Robotiq-2F-140 and suction grippers are included in the Appendix. \ourmodel is the most proficient method across all grippers, though performance varies by embodiment.
In the Franka-sim experiments, \ourmodel outperforms M2T2 by~\SI{37}{\percent}~(Fig.~\ref{fig:baseline_comparison}), with even larger margins in Robotiq-sim (\SI{44}{\percent}) and real-robot experiments (\SI{57}{\percent}).
This is likely because M2T2 relies on a contact point representation~\citep{sundermeyer2021-contact-graspnet}, which is designed for symmetric, non-adaptive grippers and struggles with adaptive grippers like the Robotiq-2F-140.
\ourmodel also outperforms SE3-Diff~\citep{urain2022se3dif} across all three grippers.

\begin{figure*}[t]
 \centering
 \includegraphics[width=\linewidth]{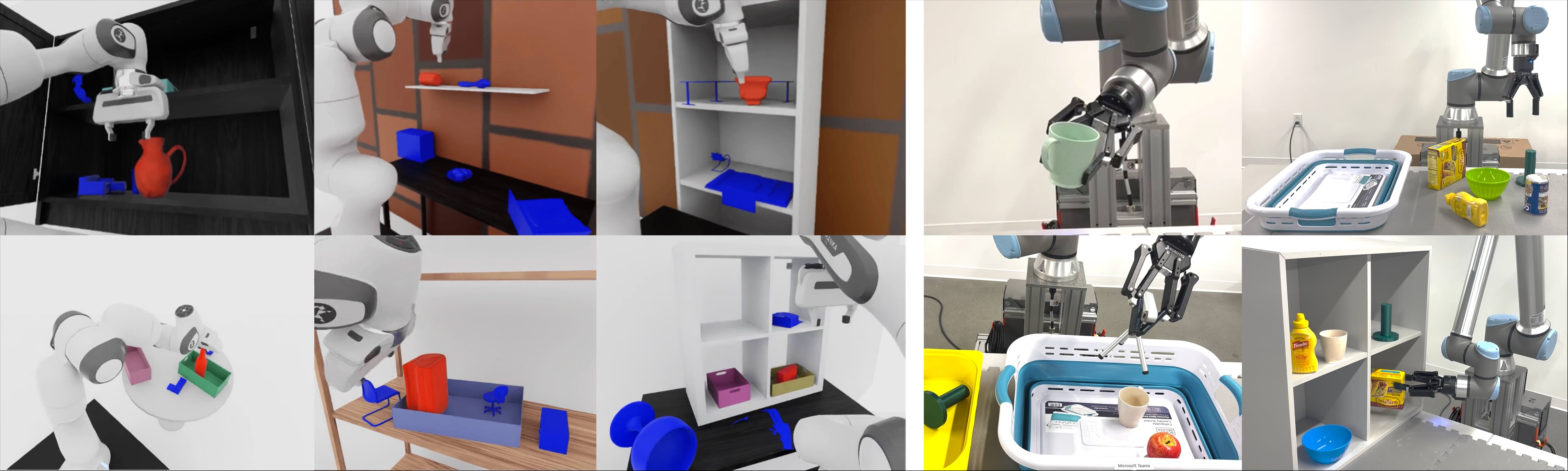}
\vspace{-1em}
 \caption{We evaluate in diverse cluttered environments in simulation \citep{han2024fetchbench} \textit{(left)} and real \textit{(right)}.}
 \label{fig:fetchbench_and_real_envs}
  \vspace{-5mm}
\end{figure*}

\subsection{Real Robot Evaluation}
\label{sec:realrobot}

We show that \ourmodel generalizes to the real world despite being only trained in simulation.
Our hardware setup consists of a UR10 arm with a single extrinsically calibrated RealSense D435 RGB-D camera overlooking a tabletop scene.
Motion planning is done with cuRobo~\citep{curobo_report23} on a Jetson while NVBlox~\citep{millane2024nvblox} is used for collision avoidance.
We use SAM2~\citep{ravi2024sam2} running on a 6000~Ada GPU for instance segmentation, as well as FoundationStereo~\citep{wen2025stereo} for depth estimation. 

We specifically evaluate the model trained on the \ourmodel Robotiq-2F-140 dataset.
We compare \ourmodel to M2T2~\citep{yuan2023m2t2} since it had the best performance in simulation among all competitors. AnyGrasp~\citep{fang2023anygrasp} is another recent grasping in clutter framework trained on real colored point clouds of tabletop objects. Due to license restrictions, we were unable to compare to AnyGrasp in our simulation experiments but evaluated it in the real world. For both methods, we use the weights and configuration released in the respective papers. We had two modifications for M2T2: a 90\degree rotation of the point cloud around \textit{z} and a 3D box crop encompassing the robot's workspace to match its training distribution. For AnyGrasp, we had to apply a translation offset along the camera's \textit{z} axis to match the original training dataset, which was collected at a fixed camera depth (despite randomized elevation/azimuth). We empirically found that inference without non-maximal suppression was better, most likely since our motion planner \citep{curobo_report23} is proficient with goal set targets. We were unable to get consistent grasp predictions without these modifications for both models. All models return a set of predicted grasps and confidence scores. We use the top-100 grasps as pose targets for the motion planner. The planner filters out grasps that are in collision or do not have an inverse kinematics solution.

\begin{wraptable}{r}{0.55\textwidth}
\vspace{-10pt}
\scriptsize
\centering
\begin{tabular}{lccccc}
\toprule
\multirow{2}{*}{\textbf{Method}} & \multirow{2}{*}{\shortstack[c]{\textbf{Isolated} \\ \textbf{Objects}}} & \multicolumn{3}{c}{\textbf{Clutter}} & \multirow{2}{*}{\textbf{Overall}} \\
\cmidrule(lr){3-5}
& & \textbf{Table} & \textbf{Basket} & \textbf{Shelf} & \\
\midrule
\ourmodel & \textbf{90.5\%} & \textbf{83.3\%} & \textbf{80.0\%} & \textbf{71.4\%} & \textbf{81.3\%} \\
M2T2 \citep{yuan2023m2t2} & 81.0\% & 75.0\% & 40.0\% & 14.3\% & 52.6\% \\
AnyGrasp \citep{fang2023anygrasp} & 85.7\% & 83.3\% & 42.9\% & 42.9\% & 63.7 \\
\bottomrule
\end{tabular}
\captionsetup{justification=centering}
\caption{Real Robot - Grasp Success Rates}
\label{tb:robot_v2}
\end{wraptable}

We evaluate four different settings: isolated objects without any clutter, multiple objects on a table, inside a basket, and on a shelf (clockwise, Fig.~\ref{fig:fetchbench_and_real_envs} \textit{(right)}).
As shown in Table~\ref{tb:robot_v2}, \ourmodel achieved an overall success rate of \SI{81.3}{\percent}, outperforming M2T2 and AnyGrasp by \SI{28}{\percent} and \SI{17.6}{\percent} respectively. \ourmodel performed well across different environments, though it struggled in the more challenging shelf and basket setting.
Motion planning is more diffucult in these settings as cuRobo filters out most grasps due to kinematic/collision restrictions. 
As such, the models need to both 1) generalize to these settings and 2) generate grasps with high coverage, to increase the chances of having feasible grasps after all the filtration steps.
Since both M2T2 and AnyGrasp are scene-centric models trained only with data for tabletop clutter, they were unable to generalize to more complicated environments.
M2T2 also did not generate grasps on some smaller objects, most likely due to the low point cloud resolution on these objects when reasoning at the scene level. More examples of grasp predictions are provided in the Appendix.

\section{Conclusion \& Limitations}

We presented \ourmodel, a 6-DOF grasp generation framework with an improved diffusion model, validated across multiple objects and three gripper embodiments.
\ourmodel outperforms baseline methods and achieves state-of-the-art results on the FetchBench~\citep{han2024fetchbench} benchmark for grasping in clutter.
We hope this framework provides a foundation for future downstream applications.

The performance of \ourmodel depends on the quality of depth sensing and instance segmentation. We noticed that \ourmodel struggled to predict grasps for cuboids in practice - we believe that training on more box-like data (which we aim to do in a next version) would resolve this. Additionally, it is computationally demanding, requiring approximately 3K GPU hours on NVIDIA V100 8-GPU nodes for data generation and training.

\clearpage
\bibliography{references}

\begin{thebibliography}{74}
\providecommand{\natexlab}[1]{#1}
\providecommand{\url}[1]{\texttt{#1}}
\expandafter\ifx\csname urlstyle\endcsname\relax
  \providecommand{\doi}[1]{doi: #1}\else
  \providecommand{\doi}{doi: \begingroup \urlstyle{rm}\Url}\fi

\bibitem[ccb()]{ccby}
Creative commons by.
\newblock URL \url{https://creativecommons.org/licenses/by/4.0/}.

\bibitem[Breyer et~al.(2020)Breyer, Chung, Ott, Roland, and
  Juan]{breyer2020volumetric}
Michel Breyer, Jen~Jen Chung, Lionel Ott, Siegwart Roland, and Nieto Juan.
\newblock Volumetric grasping network: Real-time 6 dof grasp detection in
  clutter.
\newblock In \emph{Conference on Robot Learning}, 2020.

\bibitem[Calandra et~al.(2018)Calandra, Owens, Jayaraman, Lin, Yuan, Malik,
  Adelson, and Levine]{calandra2018more}
Roberto Calandra, Andrew Owens, Dinesh Jayaraman, Justin Lin, Wenzhen Yuan,
  Jitendra Malik, Edward~H. Adelson, and Sergey Levine.
\newblock More than a feeling: Learning to grasp and regrasp using vision and
  touch.
\newblock \emph{IEEE Robotics and Automation Letters}, 3\penalty0 (4):\penalty0
  3300--3307, 2018.
\newblock \doi{10.1109/LRA.2018.2852779}.
\newblock URL \url{https://arxiv.org/abs/1805.11085}.

\bibitem[Carvalho et~al.(2024)Carvalho, Le, Jahr, Sun, Urain, Koert, and
  Peters]{carvalho2024graspdiffusionnetworklearning}
Joao Carvalho, An~T. Le, Philipp Jahr, Qiao Sun, Julen Urain, Dorothea Koert,
  and Jan Peters.
\newblock Grasp diffusion network: Learning grasp generators from partial point
  clouds with diffusion models in so(3)xr3, 2024.
\newblock URL \url{https://arxiv.org/abs/2412.08398}.

\bibitem[Casas et~al.(2024)Casas, Khargonkar, Prabhakaran, and
  Xiang]{casas2024multigrippergraspdatasetroboticgrasping}
Luis~Felipe Casas, Ninad Khargonkar, Balakrishnan Prabhakaran, and Yu~Xiang.
\newblock Multigrippergrasp: A dataset for robotic grasping from parallel jaw
  grippers to dexterous hands, 2024.
\newblock URL \url{https://arxiv.org/abs/2403.09841}.

\bibitem[Chao et~al.(2021)Chao, Yang, Xiang, Molchanov, Handa, Tremblay,
  Narang, Van~Wyk, Iqbal, Birchfield, et~al.]{chao2021dexycb}
Yu-Wei Chao, Wei Yang, Yu~Xiang, Pavlo Molchanov, Ankur Handa, Jonathan
  Tremblay, Yashraj~S Narang, Karl Van~Wyk, Umar Iqbal, Stan Birchfield, et~al.
\newblock Dexycb: A benchmark for capturing hand grasping of objects.
\newblock In \emph{Proceedings of the IEEE/CVF Conference on Computer Vision
  and Pattern Recognition}, pages 9044--9053, 2021.

\bibitem[Chen et~al.(2024{\natexlab{a}})Chen, Xie, Tang, Hu, Dai, and
  Wang]{chen2024regionawaregraspframeworknormalized}
Siang Chen, Pengwei Xie, Wei Tang, Dingchang Hu, Yixiang Dai, and Guijin Wang.
\newblock Region-aware grasp framework with normalized grasp space for
  efficient 6-dof grasping, 2024{\natexlab{a}}.
\newblock URL \url{https://arxiv.org/abs/2406.01767}.

\bibitem[Chen et~al.(2024{\natexlab{b}})Chen, Walsman, Memmel, Mo, Fang,
  Vemuri, Wu, Fox, and Gupta]{chen2024urdformer}
Zoey Chen, Aaron Walsman, Marius Memmel, Kaichun Mo, Alex Fang, Karthikeya
  Vemuri, Alan Wu, Dieter Fox, and Abhishek Gupta.
\newblock Urdformer: A pipeline for constructing articulated simulation
  environments from real-world images.
\newblock \emph{arXiv preprint arXiv:2405.11656}, 2024{\natexlab{b}}.

\bibitem[Chi et~al.(2024)Chi, Xu, Feng, Cousineau, Du, Burchfiel, Tedrake, and
  Song]{chi2024diffusionpolicyvisuomotorpolicy}
Cheng Chi, Zhenjia Xu, Siyuan Feng, Eric Cousineau, Yilun Du, Benjamin
  Burchfiel, Russ Tedrake, and Shuran Song.
\newblock Diffusion policy: Visuomotor policy learning via action diffusion,
  2024.
\newblock URL \url{https://arxiv.org/abs/2303.04137}.

\bibitem[Dalal et~al.(2024)Dalal, Liu, Talbott, Chen, Pathak, Zhang, and
  Salakhutdinov]{dalal2024manipgen}
Murtaza Dalal, Min Liu, Walter Talbott, Chen Chen, Deepak Pathak, Jian Zhang,
  and Ruslan Salakhutdinov.
\newblock Local policies enable zero-shot long-horizon manipulation.
\newblock \emph{arXiv preprint arXiv:2410.22332}, 2024.

\bibitem[Deitke et~al.(2023)Deitke, Schwenk, Salvador, Weihs, Michel,
  VanderBilt, Schmidt, Ehsani, Kembhavi, and Farhadi]{deitke2023objaverse}
Matt Deitke, Dustin Schwenk, Jordi Salvador, Luca Weihs, Oscar Michel, Eli
  VanderBilt, Ludwig Schmidt, Kiana Ehsani, Aniruddha Kembhavi, and Ali
  Farhadi.
\newblock Objaverse: A universe of annotated 3d objects.
\newblock In \emph{Proceedings of the IEEE/CVF Conference on Computer Vision
  and Pattern Recognition}, pages 13142--13153, 2023.

\bibitem[Deng et~al.(2020)Deng, Xiang, Mousavian, Eppner, Bretl, and
  Fox]{deng2020self}
Xinke Deng, Yu~Xiang, Arsalan Mousavian, Clemens Eppner, Timothy Bretl, and
  Dieter Fox.
\newblock Self-supervised 6d object pose estimation for robot manipulation.
\newblock In \emph{Proceedings of the IEEE International Conference on Robotics
  and Automation (ICRA)}, 2020.
\newblock URL \url{https://arxiv.org/abs/1909.11652}.

\bibitem[Deshpande et~al.(2025)Deshpande, Deng, Ray, Salvador, Han, Duan, Zeng,
  Zhu, Krishna, and Hendrix]{deshpande2025graspmolmo}
Abhay Deshpande, Yuquan Deng, Arijit Ray, Jordi Salvador, Winson Han, Jiafei
  Duan, Kuo-Hao Zeng, Yuke Zhu, Ranjay Krishna, and Rose Hendrix.
\newblock Graspmolmo: Generalizable task-oriented grasping via large-scale
  synthetic data generation, 2025.
\newblock URL \url{https://arxiv.org/abs/2505.13441}.

\bibitem[Eppner et~al.(2021)Eppner, Mousavian, and Fox]{eppner2021acronym}
Clemens Eppner, Arsalan Mousavian, and Dieter Fox.
\newblock Acronym: A large-scale grasp dataset based on simulation.
\newblock In \emph{2021 IEEE International Conference on Robotics and
  Automation (ICRA)}, pages 6222--6227. IEEE, 2021.

\bibitem[Eppner et~al.(2024)Eppner, Murali, Garrett, O'Flaherty, Hermans, Yang,
  and Fox]{Eppner2024}
Clemens Eppner, Adithyavairavan Murali, Caelan Garrett, Rowland O'Flaherty,
  Tucker Hermans, Wei Yang, and Dieter Fox.
\newblock scene synthesizer: A python library for procedural scene generation
  in robot manipulation.
\newblock \emph{Journal of Open Source Software}, 2024.

\bibitem[Fang et~al.(2023)Fang, Wang, Fang, Gou, Liu, Yan, Liu, Xie, and
  Lu]{fang2023anygrasp}
Hao-Shu Fang, Chenxi Wang, Hongjie Fang, Minghao Gou, Jirong Liu, Hengxu Yan,
  Wenhai Liu, Yichen Xie, and Cewu Lu.
\newblock Anygrasp: Robust and efficient grasp perception in spatial and
  temporal domains.
\newblock \emph{IEEE Transactions on Robotics (T-RO)}, 2023.

\bibitem[Fang et~al.(2020)Fang, Zhu, Garg, Kurenkov, Mehta, Fei-Fei, and
  Savarese]{fang2020learning}
Kuan Fang, Yuke Zhu, Animesh Garg, Andrey Kurenkov, Viraj Mehta, Li~Fei-Fei,
  and Silvio Savarese.
\newblock Learning task-oriented grasping for tool manipulation from simulated
  self-supervision.
\newblock \emph{The International Journal of Robotics Research}, 39\penalty0
  (2-3):\penalty0 202--216, 2020.

\bibitem[Ferrari et~al.(1992)Ferrari, Canny, et~al.]{ferrari1992planning-Q1}
Carlo Ferrari, John~F Canny, et~al.
\newblock Planning optimal grasps.
\newblock In \emph{International Conference on Robotics and Automation (ICRA)},
  volume~3, page~6. IEEE, 1992.

\bibitem[Freiberg et~al.(2024)Freiberg, Qualmann, Vien, and
  Neumann]{freiberg2024diffusionmultiembodimentgrasping}
Roman Freiberg, Alexander Qualmann, Ngo~Anh Vien, and Gerhard Neumann.
\newblock Diffusion for multi-embodiment grasping, 2024.
\newblock URL \url{https://arxiv.org/abs/2410.18835}.

\bibitem[Gupta et~al.(2019)Gupta, Dollar, and Girshick]{gupta2019lvis}
Agrim Gupta, Piotr Dollar, and Ross Girshick.
\newblock Lvis: A dataset for large vocabulary instance segmentation.
\newblock In \emph{Proceedings of the IEEE/CVF conference on computer vision
  and pattern recognition}, pages 5356--5364, 2019.

\bibitem[Hampali et~al.(2020)Hampali, Rad, Oberweger, and
  Lepetit]{hampali2020honnotate-ho3d}
Shreyas Hampali, Mahdi Rad, Markus Oberweger, and Vincent Lepetit.
\newblock Honnotate: A method for 3d annotation of hand and object poses.
\newblock In \emph{Proceedings of the IEEE/CVF conference on computer vision
  and pattern recognition}, pages 3196--3206, 2020.

\bibitem[Han et~al.(2024)Han, Parakh, Geng, Defay, Gan, and
  Deng]{han2024fetchbench}
Beining Han, Meenal Parakh, Derek Geng, Jack~A Defay, Luyang Gan, and Jia Deng.
\newblock Fetchbench: A simulation benchmark for robot fetching.
\newblock \emph{arXiv preprint arXiv:2406.11793}, 2024.

\bibitem[Ho et~al.(2020)Ho, Jain, and Abbeel]{ho2020denoising}
Jonathan Ho, Ajay Jain, and Pieter Abbeel.
\newblock Denoising diffusion probabilistic models.
\newblock \emph{arXiv preprint arxiv:2006.11239}, 2020.

\bibitem[Huang et~al.(2024{\natexlab{a}})Huang, Sundaralingam, Mousavian,
  Murali, Goldberg, and Fox]{huangdiffusionseeder}
Huang Huang, Balakumar Sundaralingam, Arsalan Mousavian, Adithyavairavan
  Murali, Ken Goldberg, and Dieter Fox.
\newblock Diffusionseeder: Seeding motion optimization with diffusion for rapid
  motion planning.
\newblock 2024{\natexlab{a}}.

\bibitem[Huang et~al.(2024{\natexlab{b}})Huang, Wang, Li, Zhang, and
  Fei-Fei]{huang2024rekep}
Wenlong Huang, Chen Wang, Yunzhu Li, Ruohan Zhang, and Li~Fei-Fei.
\newblock Rekep: Spatio-temporal reasoning of relational keypoint constraints
  for robotic manipulation.
\newblock \emph{arXiv preprint arXiv:2409.01652}, 2024{\natexlab{b}}.

\bibitem[Jiang et~al.(2021)Jiang, Zhu, Svetlik, Fang, and
  Zhu]{jiang2021synergies}
Zhenyu Jiang, Yifeng Zhu, Maxwell Svetlik, Kuan Fang, and Yuke Zhu.
\newblock Synergies between affordance and geometry: 6-dof grasp detection via
  implicit representations.
\newblock \emph{Robotics: science and systems}, 2021.

\bibitem[Ju et~al.(2025)Ju, Hu, Zhang, Zhang, Jiang, and Xu]{ju2025robo}
Yuanchen Ju, Kaizhe Hu, Guowei Zhang, Gu~Zhang, Mingrun Jiang, and Huazhe Xu.
\newblock Robo-abc: Affordance generalization beyond categories via semantic
  correspondence for robot manipulation.
\newblock In \emph{European Conference on Computer Vision}, pages 222--239.
  Springer, 2025.

\bibitem[Kalashnikov et~al.(2018)Kalashnikov, Irpan, Pastor, Ibarz, Herzog,
  Jang, Quillen, Holly, Kalakrishnan, Vanhoucke, and Levine]{qtopt2018}
Dmitry Kalashnikov, Alex Irpan, Peter Pastor, Julian Ibarz, Alexander Herzog,
  Eric Jang, Deirdre Quillen, Ethan Holly, Mrinal Kalakrishnan, Vincent
  Vanhoucke, and Sergey Levine.
\newblock Qt-opt: Scalable deep reinforcement learning for vision-based robotic
  manipulation.
\newblock \emph{Conference on Robot Learning}, 2018.

\bibitem[Ke et~al.(2024)Ke, Gkanatsios, and
  Fragkiadaki]{ke20243ddiffuseractorpolicy}
Tsung-Wei Ke, Nikolaos Gkanatsios, and Katerina Fragkiadaki.
\newblock 3d diffuser actor: Policy diffusion with 3d scene representations,
  2024.
\newblock URL \url{https://arxiv.org/abs/2402.10885}.

\bibitem[Kuang et~al.(2024)Kuang, Ye, Geng, Mao, Deng, Guibas, Wang, and
  Wang]{kuang2024ram}
Yuxuan Kuang, Junjie Ye, Haoran Geng, Jiageng Mao, Congyue Deng, Leonidas
  Guibas, He~Wang, and Yue Wang.
\newblock Ram: Retrieval-based affordance transfer for generalizable zero-shot
  robotic manipulation.
\newblock \emph{arXiv preprint arXiv:2407.04689}, 2024.

\bibitem[Li et~al.(2023)Li, Liu, Li, Geng, Zhu, Yang, and
  Huang]{li2023gendexgrasp}
Puhao Li, Tengyu Liu, Yuyang Li, Yiran Geng, Yixin Zhu, Yaodong Yang, and
  Siyuan Huang.
\newblock Gendexgrasp: Generalizable dexterous grasping.
\newblock In \emph{2023 IEEE International Conference on Robotics and
  Automation (ICRA)}, pages 8068--8074. IEEE, 2023.

\bibitem[Liang et~al.(2019)Liang, Ma, Li, Gorner, Tang, Fang, Sun, and
  Zhang]{PointNetGPD2019}
Hongzhuo Liang, Xiaojian Ma, Shuang Li, Michael Gorner, Song Tang, Bin Fang,
  Fuchun Sun, and Jianwei Zhang.
\newblock Pointnetgpd: Detecting grasp configurations from point sets.
\newblock In \emph{IEEE International Conference on Robotics and Automation},
  2019.

\bibitem[Liu et~al.(2020)Liu, Pan, Xu, Ganguly, and Manocha]{liu2020deep-ddg}
Min Liu, Zherong Pan, Kai Xu, Kanishka Ganguly, and Dinesh Manocha.
\newblock Deep differentiable grasp planner for high-dof grippers.
\newblock \emph{arXiv preprint arXiv:2002.01530}, 2020.

\bibitem[Liu et~al.(2024)Liu, Orru, Paxton, Shafiullah, and
  Pinto]{liu2024okrobot}
Peiqi Liu, Yaswanth Orru, Chris Paxton, Nur Muhammad~Mahi Shafiullah, and
  Lerrel Pinto.
\newblock Ok-robot: What really matters in integrating open-knowledge models
  for robotics.
\newblock \emph{arXiv preprint arXiv:2401.12202}, 2024.

\bibitem[Liu et~al.(2023)Liu, Du, Hermans, Chernova, and
  Paxton]{structdiffusion2023}
Weiyu Liu, Yilun Du, Tucker Hermans, Sonia Chernova, and Chris Paxton.
\newblock Structdiffusion: Language-guided creation of physically-valid
  structures using unseen objects.
\newblock In \emph{RSS 2023}, 2023.

\bibitem[Lum et~al.(2024)Lum, Li, Culbertson, Srinivasan, Ames, Schwager, and
  Bohg]{lum2024get}
Tyler Ga~Wei Lum, Albert~H. Li, Preston Culbertson, Krishnan Srinivasan, Aaron
  Ames, Mac Schwager, and Jeannette Bohg.
\newblock Get a grip: Multi-finger grasp evaluation at scale enables robust
  sim-to-real transfer.
\newblock In \emph{8th Annual Conference on Robot Learning}, 2024.
\newblock URL \url{https://openreview.net/forum?id=1jc2zA5Z6J}.

\bibitem[Macklin et~al.(2014)Macklin, M{\"u}ller, Chentanez, and
  Kim]{macklin2014unified-flex-simulator}
Miles Macklin, Matthias M{\"u}ller, Nuttapong Chentanez, and Tae-Yong Kim.
\newblock Unified particle physics for real-time applications.
\newblock \emph{ACM Transactions on Graphics (TOG)}, 33\penalty0 (4):\penalty0
  1--12, 2014.

\bibitem[Mahler et~al.(2018)Mahler, Matl, Liu, Li, Gealy, and
  Goldberg]{mahler2018dexnet30computingrobust}
Jeffrey Mahler, Matthew Matl, Xinyu Liu, Albert Li, David Gealy, and Ken
  Goldberg.
\newblock Dex-net 3.0: Computing robust robot vacuum suction grasp targets in
  point clouds using a new analytic model and deep learning, 2018.
\newblock URL \url{https://arxiv.org/abs/1709.06670}.

\bibitem[Makoviychuk et~al.(2021)Makoviychuk, Wawrzyniak, Guo, Lu, Storey,
  Macklin, Hoeller, Rudin, Allshire, Handa, and State]{issacgym2021}
Viktor Makoviychuk, Lukasz Wawrzyniak, Yunrong Guo, Michelle Lu, Kier Storey,
  Miles Macklin, David Hoeller, Nikita Rudin, Arthur Allshire, Ankur Handa, and
  Gavriel State.
\newblock Isaac gym: High performance gpu-based physics simulation for robot
  learning.
\newblock \emph{arXiv:2108.10470}, 2021.

\bibitem[Millane et~al.(2024)Millane, Oleynikova, Wirbel, Steiner, Ramasamy,
  Tingdahl, and Siegwart]{millane2024nvblox}
Alexander Millane, Helen Oleynikova, Emilie Wirbel, Remo Steiner, Vikram
  Ramasamy, David Tingdahl, and Roland Siegwart.
\newblock nvblox: Gpu-accelerated incremental signed distance field mapping,
  2024.

\bibitem[Morrison et~al.(2020)Morrison, Corke, and Leitner]{morrison2020egad}
Douglas Morrison, Peter Corke, and J{\"u}rgen Leitner.
\newblock Egad! an evolved grasping analysis dataset for diversity and
  reproducibility in robotic manipulation.
\newblock \emph{IEEE Robotics and Automation Letters}, 5\penalty0 (3):\penalty0
  4368--4375, 2020.

\bibitem[Mousavian et~al.(2019)Mousavian, Eppner, and
  Fox]{mousavian2019-6dofgraspnet}
Arsalan Mousavian, Clemens Eppner, and Dieter Fox.
\newblock 6-dof graspnet: Variational grasp generation for object manipulation.
\newblock In \emph{Proceedings of the IEEE/CVF International Conference on
  Computer Vision}, pages 2901--2910, 2019.

\bibitem[Murali et~al.(2018)Murali, Li, Gandhi, and Gupta]{murali2018iser}
Adithyavairavan Murali, Yin Li, Dhiraj Gandhi, and Abhinav Gupta.
\newblock Learning to grasp without seeing.
\newblock In \emph{Proceedings of the 2018 International Symposium on
  Experimental Robotics (ISER)}, pages 375--386. Springer, 2018.
\newblock \doi{10.1007/978-3-030-33950-0_33}.
\newblock URL
  \url{https://link.springer.com/chapter/10.1007/978-3-030-33950-0_33}.

\bibitem[Murali et~al.(2020{\natexlab{a}})Murali, Liu, Marino, Chernova, and
  Gupta]{murali2020taskgrasp}
Adithyavairavan Murali, Weiyu Liu, Kenneth Marino, Sonia Chernova, and Abhinav
  Gupta.
\newblock Same object, different grasps: Data and semantic knowledge for
  task-oriented grasping.
\newblock In \emph{Conference on Robot Learning}, 2020{\natexlab{a}}.

\bibitem[Murali et~al.(2020{\natexlab{b}})Murali, Mousavian, Eppner, Paxton,
  and Fox]{Murali2020CollisionNet}
Adithyavairavan Murali, Arsalan Mousavian, Clemens Eppner, Chris Paxton, and
  Dieter Fox.
\newblock 6-dof grasping for target-driven object manipulation in clutter.
\newblock In \emph{IEEE International Conference on Robotics and Automation
  (ICRA)}, 2020{\natexlab{b}}.

\bibitem[Newbury et~al.(2022)Newbury, Gu, Chumbley, Mousavian, Eppner, Leitner,
  Bohg, Morales, Asfour, Kragic, Fox, and Cosgun]{newbury2022review}
Rhys Newbury, Morris Gu, Lachlan Chumbley, Arsalan Mousavian, Clemens Eppner,
  Jürgen Leitner, Jeannette Bohg, Antonio Morales, Tamim Asfour, Danica
  Kragic, Dieter Fox, and Akansel Cosgun.
\newblock Deep learning approaches to grasp synthesis: A review, 2022.

\bibitem[NVIDIA(2023)]{nvidia2023-isaac-sim}
NVIDIA.
\newblock Nvidia isaac sim: Robotics simulation and synthetic data, 2023.
\newblock URL \url{https://developer.nvidia.com/isaac-sim}.

\bibitem[Pokorny and Kragic(2013)]{pokorny2013classical-fastgrasp}
Florian~T Pokorny and Danica Kragic.
\newblock Classical grasp quality evaluation: New algorithms and theory.
\newblock In \emph{2013 IEEE/RSJ International Conference on Intelligent Robots
  and Systems}, pages 3493--3500. IEEE, 2013.

\bibitem[Qi et~al.(2017)Qi, Yi, Su, and Guibas]{pointnet}
Charles~R Qi, Li~Yi, Hao Su, and Leonidas~J Guibas.
\newblock Pointnet++: Deep hierarchical feature learning on point sets in a
  metric space.
\newblock \emph{Neural Information Processing Systems (NeurIPS)}, 2017.

\bibitem[Ravi et~al.(2024)Ravi, Gabeur, Hu, Hu, Ryali, Ma, Khedr, R{\"a}dle,
  Rolland, Gustafson, Mintun, Pan, Alwala, Carion, Wu, Girshick, Doll{\'a}r,
  and Feichtenhofer]{ravi2024sam2}
Nikhila Ravi, Valentin Gabeur, Yuan-Ting Hu, Ronghang Hu, Chaitanya Ryali,
  Tengyu Ma, Haitham Khedr, Roman R{\"a}dle, Chloe Rolland, Laura Gustafson,
  Eric Mintun, Junting Pan, Kalyan~Vasudev Alwala, Nicolas Carion, Chao-Yuan
  Wu, Ross Girshick, Piotr Doll{\'a}r, and Christoph Feichtenhofer.
\newblock Sam 2: Segment anything in images and videos.
\newblock \emph{arXiv preprint arXiv:2408.00714}, 2024.
\newblock URL \url{https://arxiv.org/abs/2408.00714}.

\bibitem[Savva et~al.(2015)Savva, Chang, and Hanrahan]{savva2015semantically}
Manolis Savva, Angel~X Chang, and Pat Hanrahan.
\newblock Semantically-enriched 3d models for common-sense knowledge.
\newblock In \emph{Proceedings of the IEEE Conference on Computer Vision and
  Pattern Recognition Workshops}, pages 24--31, 2015.

\bibitem[Shao et~al.(2020)Shao, Ferreira, Jorda, Nambiar, Luo, Solowjow, Ojea,
  Khatib, and Bohg]{shao2020unigrasp}
Lin Shao, Fabio Ferreira, Mikael Jorda, Varun Nambiar, Jianlan Luo, Eugen
  Solowjow, Juan~Aparicio Ojea, Oussama Khatib, and Jeannette Bohg.
\newblock Unigrasp: Learning a unified model to grasp with multifingered
  robotic hands.
\newblock \emph{IEEE Robotics and Automation Letters}, 5\penalty0 (2):\penalty0
  2286--2293, 2020.

\bibitem[Song et~al.(2024)Song, Li, and Detry]{song2024implicitgraspdiffusion}
Pinhao Song, Pengteng Li, and Renaud Detry.
\newblock Implicit grasp diffusion: Bridging the gap between dense prediction
  and sampling-based grasping.
\newblock In \emph{Conference on Robot Learning}, 2024.

\bibitem[Song and Ermon(2019)]{song2019}
Yang Song and Stefano Ermon.
\newblock Generative modeling by estimating gradients of the data distribution.
\newblock \emph{In Advances in Neural Information Processing Systems}, 2019.

\bibitem[Song et~al.(2021)Song, Sohl-Dickstein, Kingma, Kumar, Ermon, and
  Poole]{song2021b}
Yang Song, Jascha Sohl-Dickstein, Diederik Kingma, Abhishek Kumar, Stefano
  Ermon, and Ben Poole.
\newblock Score based generative modeling through stochastic differential
  equations.
\newblock \emph{International Conference on Learning Representations (ICLR)},
  2021.
\newblock URL \url{https://arxiv.org/pdf/2011.13456}.

\bibitem[Sundaralingam et~al.(2023)Sundaralingam, Hari, Fishman, Garrett, Wyk,
  Blukis, Millane, Oleynikova, Handa, Ramos, Ratliff, and Fox]{curobo_report23}
Balakumar Sundaralingam, Siva Kumar~Sastry Hari, Adam Fishman, Caelan Garrett,
  Karl~Van Wyk, Valts Blukis, Alexander Millane, Helen Oleynikova, Ankur Handa,
  Fabio Ramos, Nathan Ratliff, and Dieter Fox.
\newblock curobo: Parallelized collision-free minimum-jerk robot motion
  generation, 2023.

\bibitem[Sundermeyer et~al.(2021)Sundermeyer, Mousavian, Triebel, and
  Fox]{sundermeyer2021-contact-graspnet}
Martin Sundermeyer, Arsalan Mousavian, Rudolph Triebel, and Dieter Fox.
\newblock Contact-graspnet: Efficient 6-dof grasp generation in cluttered
  scenes.
\newblock In \emph{2021 IEEE International Conference on Robotics and
  Automation (ICRA)}, pages 13438--13444. IEEE, 2021.

\bibitem[Tang et~al.(2023)Tang, Huang, Ge, Liu, and Zhang]{tang2023graspgpt}
Chao Tang, Dehao Huang, Wenqi Ge, Weiyu Liu, and Hong Zhang.
\newblock Graspgpt: Leveraging semantic knowledge from a large language model
  for task-oriented grasping.
\newblock \emph{IEEE Robotics and Automation Letters}, 2023.

\bibitem[Tang et~al.(2024)Tang, Huang, Dong, Xu, and
  Zhang]{tang2024foundationgraspgeneralizabletaskorientedgrasping}
Chao Tang, Dehao Huang, Wenlong Dong, Ruinian Xu, and Hong Zhang.
\newblock Foundationgrasp: Generalizable task-oriented grasping with foundation
  models, 2024.
\newblock URL \url{https://arxiv.org/abs/2404.10399}.

\bibitem[Tobin et~al.(2018)Tobin, Biewald, Duan, Andrychowicz, Handa, Kumar,
  McGrew, Ray, Schneider, Welinder, Zaremba, and Abbeel]{tobin2018grasp}
Josh Tobin, Lukas Biewald, Rocky Duan, Marcin Andrychowicz, Ankur Handa, Vikash
  Kumar, Bob McGrew, Alex Ray, Jonas Schneider, Peter Welinder, Wojciech
  Zaremba, and Pieter Abbeel.
\newblock Domain randomization and generative models for robotic grasping.
\newblock 2018.
\newblock URL \url{https://arxiv.org/pdf/1710.06425v2}.

\bibitem[Turpin et~al.(2023)Turpin, Zhong, Zhang, Zhu, Liu, Singh, Heiden,
  Macklin, Tsogkas, Dickinson, et~al.]{turpin2023fast-graspd}
Dylan Turpin, Tao Zhong, Shutong Zhang, Guanglei Zhu, Jingzhou Liu, Ritvik
  Singh, Eric Heiden, Miles Macklin, Stavros Tsogkas, Sven Dickinson, et~al.
\newblock Fast-grasp'd: Dexterous multi-finger grasp generation through
  differentiable simulation.
\newblock \emph{arXiv preprint arXiv:2306.08132}, 2023.

\bibitem[Urain et~al.(2023)Urain, Funk, Peters, and
  Chalvatzaki]{urain2022se3dif}
Julen Urain, Niklas Funk, Jan Peters, and Georgia Chalvatzaki.
\newblock Se(3)-diffusionfields: Learning smooth cost functions for joint grasp
  and motion optimization through diffusion.
\newblock \emph{IEEE International Conference on Robotics and Automation
  (ICRA)}, 2023.

\bibitem[Wang et~al.(2023)Wang, Zhang, Chen, Xu, Li, Liu, and
  Wang]{wang2023dexgraspnet}
Ruicheng Wang, Jialiang Zhang, Jiayi Chen, Yinzhen Xu, Puhao Li, Tengyu Liu,
  and He~Wang.
\newblock Dexgraspnet: A large-scale robotic dexterous grasp dataset for
  general objects based on simulation.
\newblock In \emph{2023 IEEE International Conference on Robotics and
  Automation (ICRA)}, pages 11359--11366. IEEE, 2023.

\bibitem[Wen et~al.(2025)Wen, Trepte, Aribido, Kautz, Gallo, and
  Birchfield]{wen2025stereo}
Bowen Wen, Matthew Trepte, Joseph Aribido, Jan Kautz, Orazio Gallo, and Stan
  Birchfield.
\newblock Foundationstereo: Zero-shot stereo matching.
\newblock \emph{arXiv}, 2025.

\bibitem[Weng et~al.(2024)Weng, Lu, Kragic, and Lundell]{weng2024dexdiffuser}
Zehang Weng, Haofei Lu, Danica Kragic, and Jens Lundell.
\newblock Dexdiffuser: Generating dexterous grasps with diffusion models.
\newblock \emph{IEEE Robotics and Automation Letters}, 9\penalty0
  (12):\penalty0 11834--11840, 2024.
\newblock \doi{10.1109/LRA.2024.3498776}.

\bibitem[Wu et~al.(2022)Wu, Lao, Jiang, Liu, and Zhao]{wu2022ptv2}
Xiaoyang Wu, Yixing Lao, Li~Jiang, Xihui Liu, and Hengshuang Zhao.
\newblock Point transformer v2: Grouped vector attention and partition-based
  pooling.
\newblock In \emph{NeurIPS}, 2022.

\bibitem[Wu et~al.(2024)Wu, Jiang, Wang, Liu, Liu, Qiao, Ouyang, He, and
  Zhao]{wu2024ptv3}
Xiaoyang Wu, Li~Jiang, Peng-Shuai Wang, Zhijian Liu, Xihui Liu, Yu~Qiao, Wanli
  Ouyang, Tong He, and Hengshuang Zhao.
\newblock Point transformer v3: Simpler, faster, stronger.
\newblock In \emph{CVPR}, 2024.

\bibitem[Wu et~al.(2023)Wu, Wang, and
  Wang]{wu2023learning-dex-human-affordance}
Yueh-Hua Wu, Jiashun Wang, and Xiaolong Wang.
\newblock Learning generalizable dexterous manipulation from human grasp
  affordance.
\newblock In \emph{Conference on Robot Learning}, pages 618--629. PMLR, 2023.

\bibitem[Xie et~al.(2024)Xie, Chen, Tang, Hu, Yang, and
  Wang]{xie2024rethinking6dofgraspdetection}
Pengwei Xie, Siang Chen, Wei Tang, Dingchang Hu, Wenming Yang, and Guijin Wang.
\newblock Rethinking 6-dof grasp detection: A flexible framework for
  high-quality grasping, 2024.
\newblock URL \url{https://arxiv.org/abs/2403.15054}.

\bibitem[Xu et~al.(2021)Xu, Qi, Agrawal, and Song]{xu2021adagrasp}
Zhenjia Xu, Beichun Qi, Shubham Agrawal, and Shuran Song.
\newblock Adagrasp: Learning an adaptive gripper-aware grasping policy.
\newblock In \emph{2021 IEEE International Conference on Robotics and
  Automation (ICRA)}, pages 4620--4626. IEEE, 2021.

\bibitem[Yuan et~al.(2023)Yuan, Murali, Mousavian, and Fox]{yuan2023m2t2}
Wentao Yuan, Adithyavairavan Murali, Arsalan Mousavian, and Dieter Fox.
\newblock M2t2: Multi-task masked transformer for object-centric pick and
  place.
\newblock In \emph{7th Annual Conference on Robot Learning}, 2023.

\bibitem[Zhang et~al.(2024)Zhang, He, Wan, Zhang, Deng, Ma, and
  Wang]{zhang2024diffgraspwholebodygraspingsynthesis}
Yonghao Zhang, Qiang He, Yanguang Wan, Yinda Zhang, Xiaoming Deng, Cuixia Ma,
  and Hongan Wang.
\newblock Diffgrasp: Whole-body grasping synthesis guided by object motion
  using a diffusion model, 2024.
\newblock URL \url{https://arxiv.org/abs/2412.20657}.

\bibitem[Zhou et~al.(2018)Zhou, Park, and Koltun]{Zhou2018}
Qian-Yi Zhou, Jaesik Park, and Vladlen Koltun.
\newblock {Open3D}: {A} modern library for {3D} data processing.
\newblock \emph{arXiv:1801.09847}, 2018.

\bibitem[Zhou et~al.(2019)Zhou, Barnes, Jingwan, Jimei, and
  Hao]{Zhou_2019_CVPR}
Yi~Zhou, Connelly Barnes, Lu~Jingwan, Yang Jimei, and Li~Hao.
\newblock On the continuity of rotation representations in neural networks.
\newblock In \emph{The IEEE Conference on Computer Vision and Pattern
  Recognition (CVPR)}, June 2019.

\end{thebibliography}

\setcitestyle{numbers}
\bibliographystyle{plainnat}

\newpage
\clearpage
\section{Appendix}

We provide additional details on the following in this supplementary material: 
\begin{itemize}[label={},left=5em]
  \item[\textbf{Sec.~\ref{supp:vizgrasps}}] qualitative visualizations of the grasp predictions,
  \item[\textbf{Sec.~\ref{supp:dataset}}] dataset statistics,
  \item[\textbf{Sec.~\ref{supp:radarchart}}] radar chart,
  \item[\textbf{Sec.~\ref{supp:robotiq}}] evaluations on the \ourmodel Robotiq-2f-140 dataset,
  \item[\textbf{Sec.~\ref{supp:suction}}] evaluations on the \ourmodel suction dataset,
  \item[\textbf{Sec.~\ref{supp:ablations}}] ablations,
  \item[\textbf{Sec.~\ref{supp:fetchbench}}] FetchBench evaluations in clutter,
  \item[\textbf{Sec.~\ref{supp:emd}}] Earth-Movers Distance calculation
  \item[\textbf{Sec.~\ref{supp:augmentation}}] importance of noise augmentation for sim2real, and
  \item[\textbf{Sec.~\ref{supp:tuning}}] a empirical guide to tuning the number of inferred grasps.
\end{itemize}
Please also see the project video for examples of the real robot execution.

\subsection{Qualitative Visualizations of Grasp Predictions}
\label{supp:vizgrasps}

\begin{figure}
    \centering
    \includegraphics[width=1\linewidth]{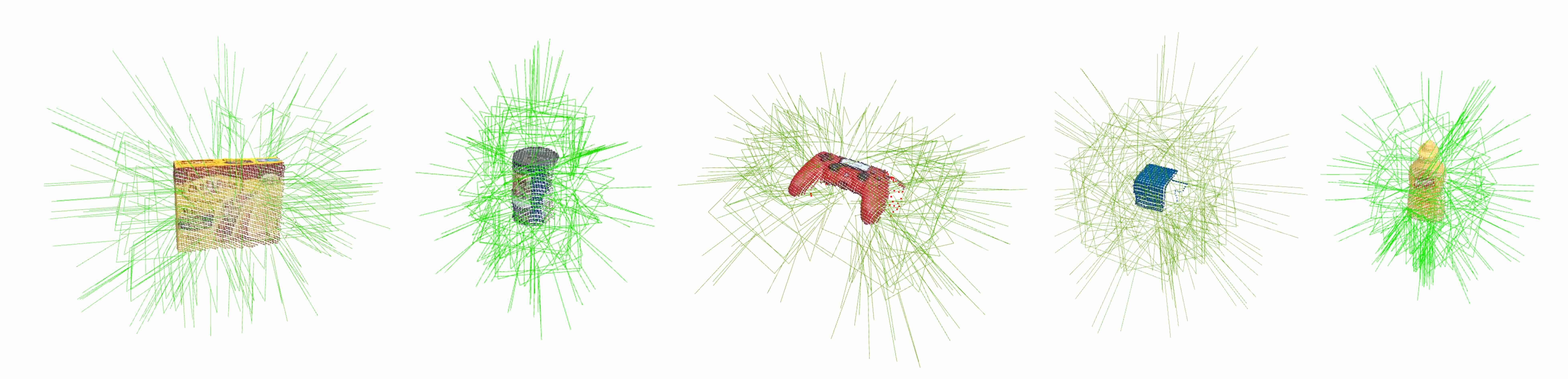}
    \caption{Examples of grasp predictions overlaid on segmental partial point clouds from real objects.}
    \label{fig:realpc}
\end{figure}

Qualitative grasp predictions are shown in Table \ref{tab:grasp_predictions}. Overall, \ourmodel's predictions are more focused on the target object and have greater coverage, when compared to the M2T2 \cite{yuan2023m2t2} baseline. M2T2 sometimes does not generate any grasps for some target objects~(the bell pepper in the 2nd row), especially when they are small. We believe this is because M2T2 is a scene-centric model, and the resolution is insufficient to capture the geometry of smaller objects (i.e. there are too few points on them) - this is unavoidable for models reasoning at the scene-level. Furthermore, M2T2 generates several false positive grasps in the environment, which may seep into the contact mask of any neighboring target objects. Additional \ourmodel grasp predictions on segmented object point clouds (from real objects) are shown in Fig \ref{fig:realpc}.

\subsection{Further Dataset Statistics}
\label{supp:dataset}
A detailed comparison between our \ourmodel dataset and prior work in the literature is shown in Table \ref{tb:related-datasets}.

\begin{table}
    \centering
    \scriptsize
    \begin{tabularx}{\linewidth}{XXX}
        \toprule
        \textbf{RGB +  Segmentation Mask} & \textbf{\ourmodel (Ours)} & \textbf{M2T2} \cite{yuan2023m2t2}  \\ \midrule
         \includegraphics[width=\linewidth]{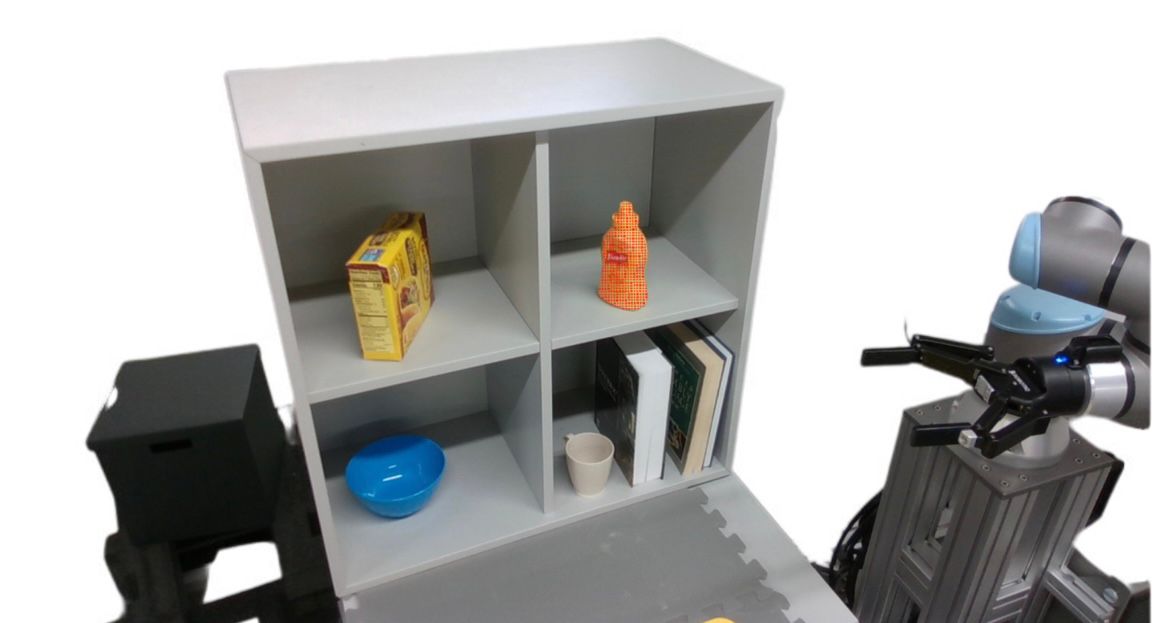} & \includegraphics[width=\linewidth]{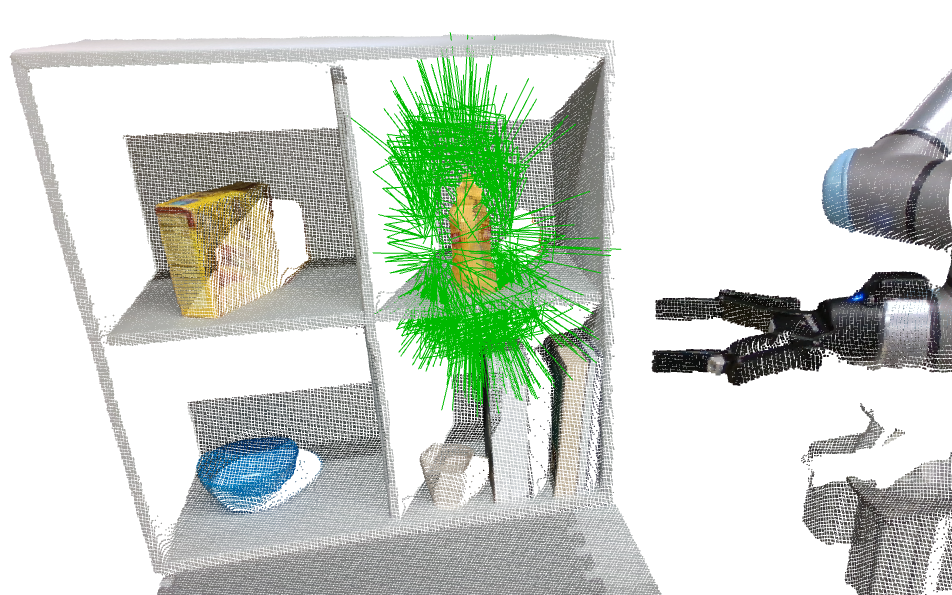} & \includegraphics[width=\linewidth]{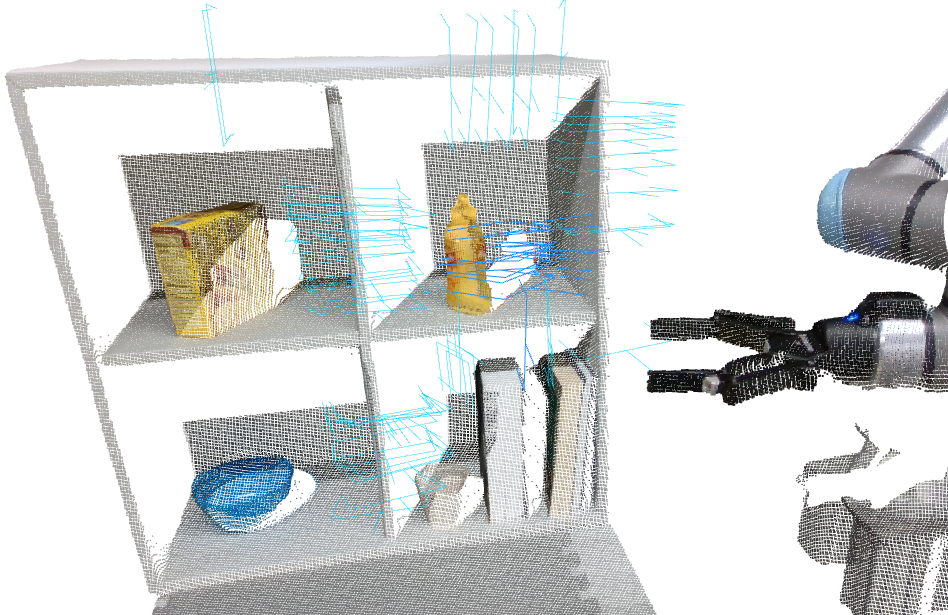} 
\\ \midrule
         \includegraphics[width=\linewidth]{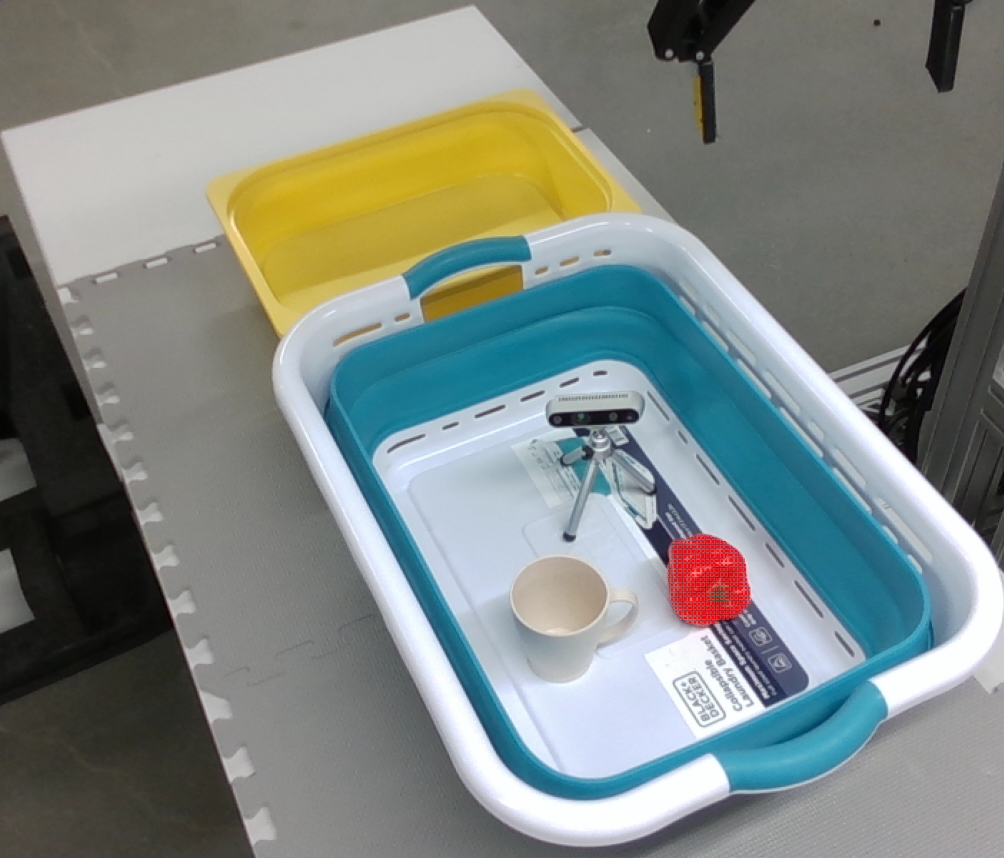} & \includegraphics[width=\linewidth]{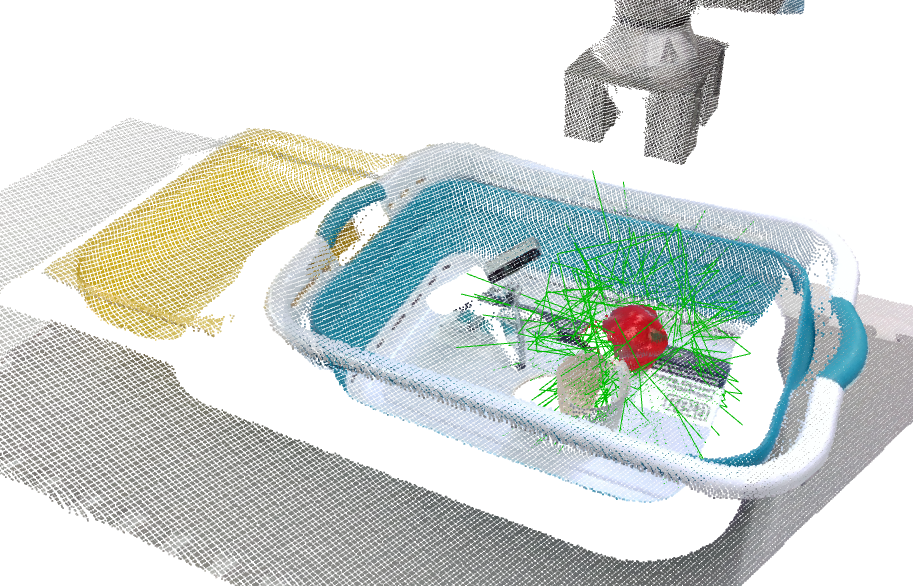} & \includegraphics[width=\linewidth]{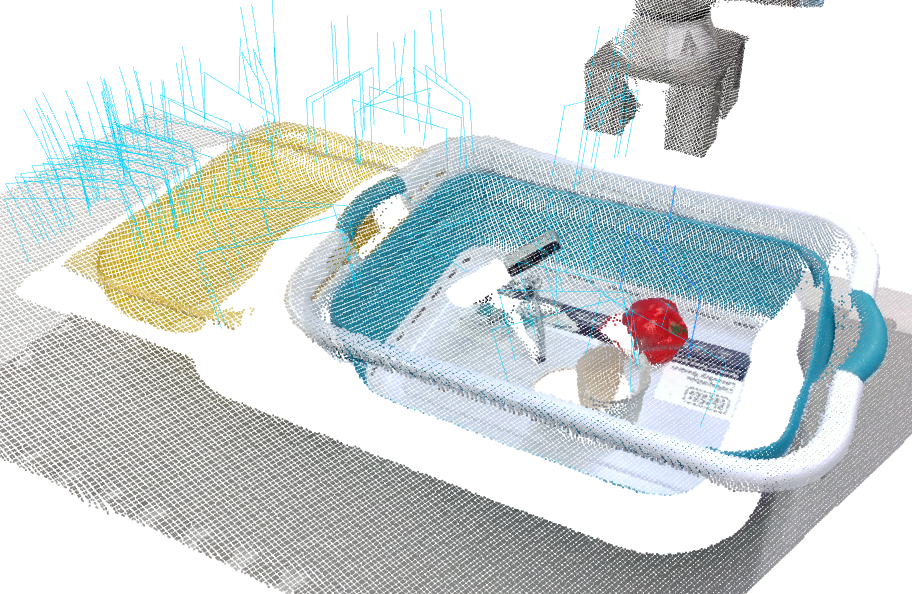} 
\\ \midrule
         \includegraphics[width=\linewidth]{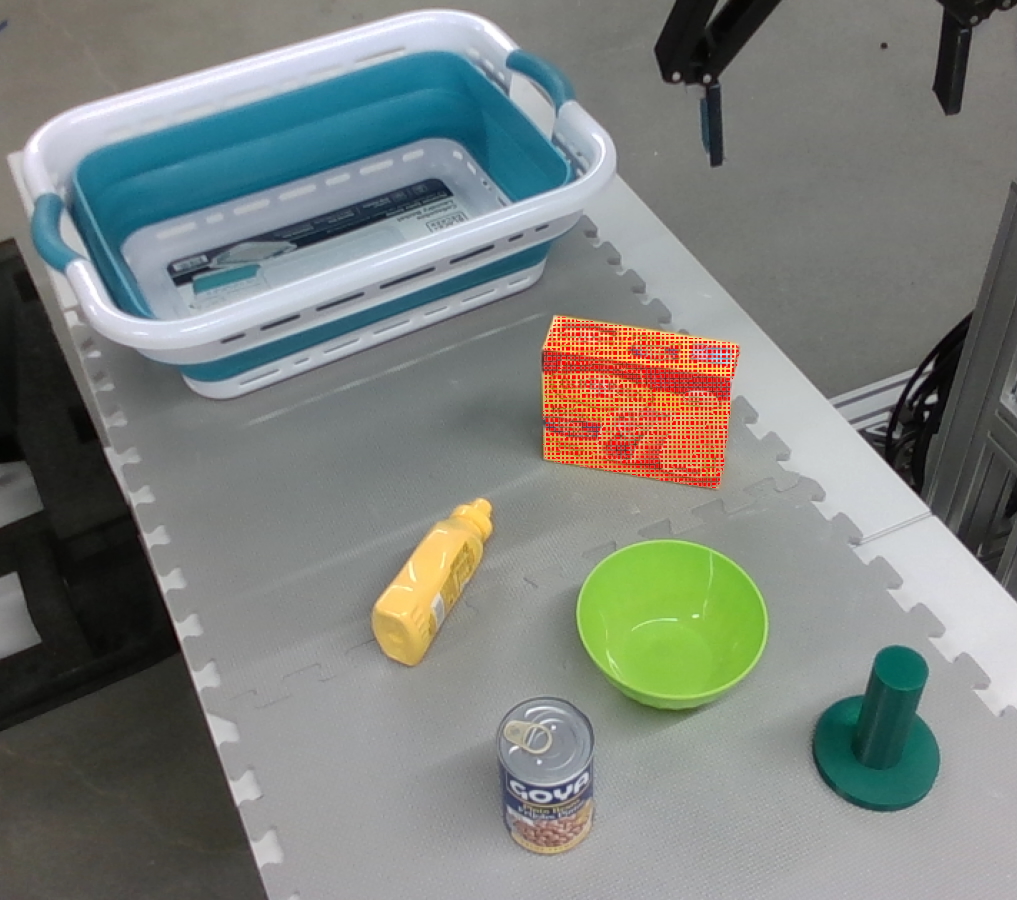} & \includegraphics[width=\linewidth]{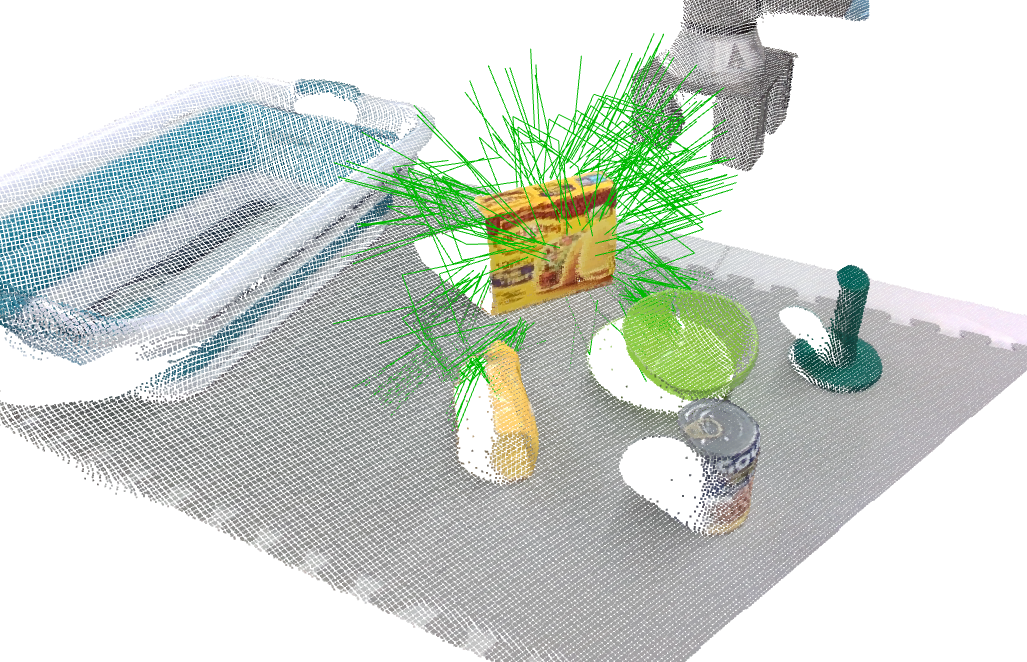} & \includegraphics[width=\linewidth]{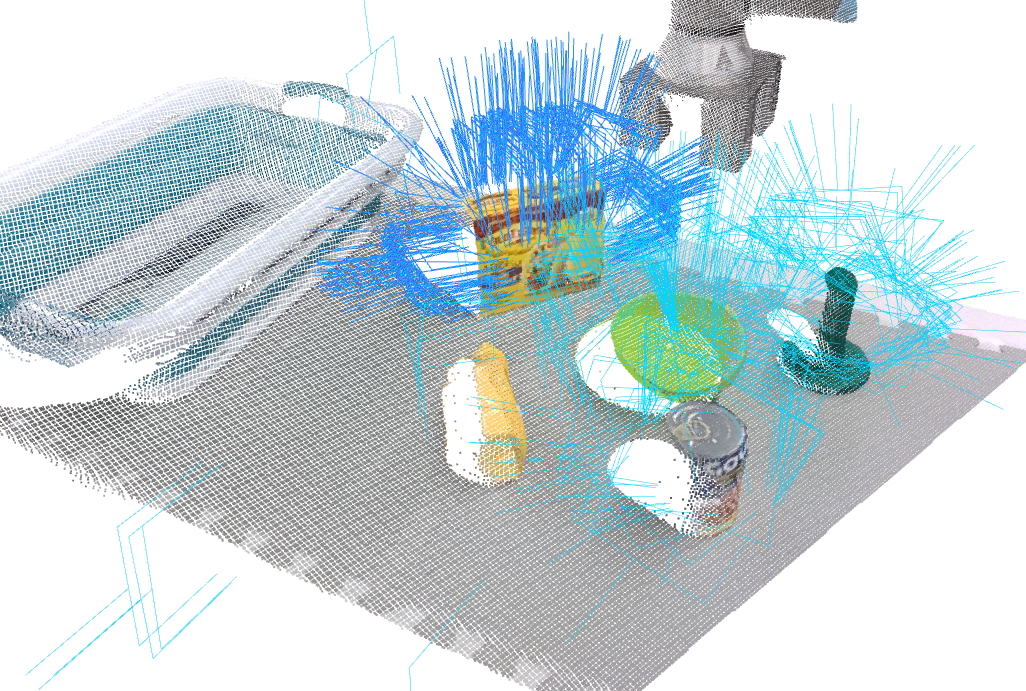} 
\\ \midrule
         \includegraphics[width=\linewidth]{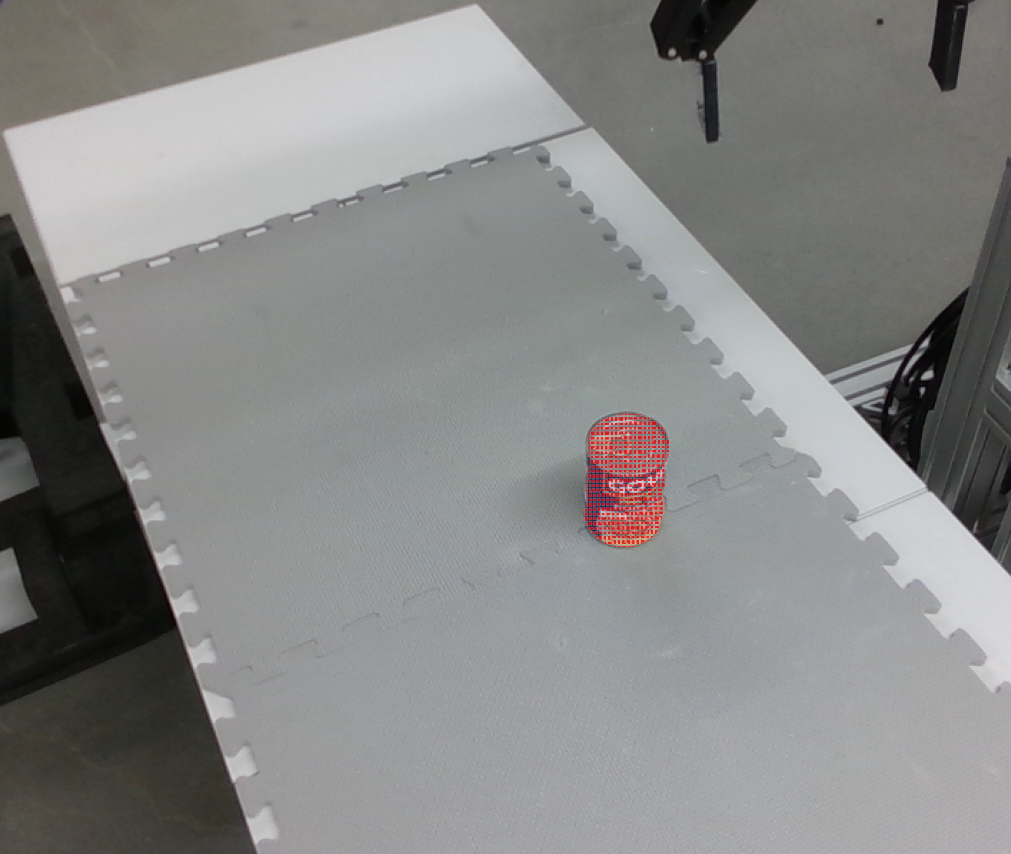} & \includegraphics[width=\linewidth]{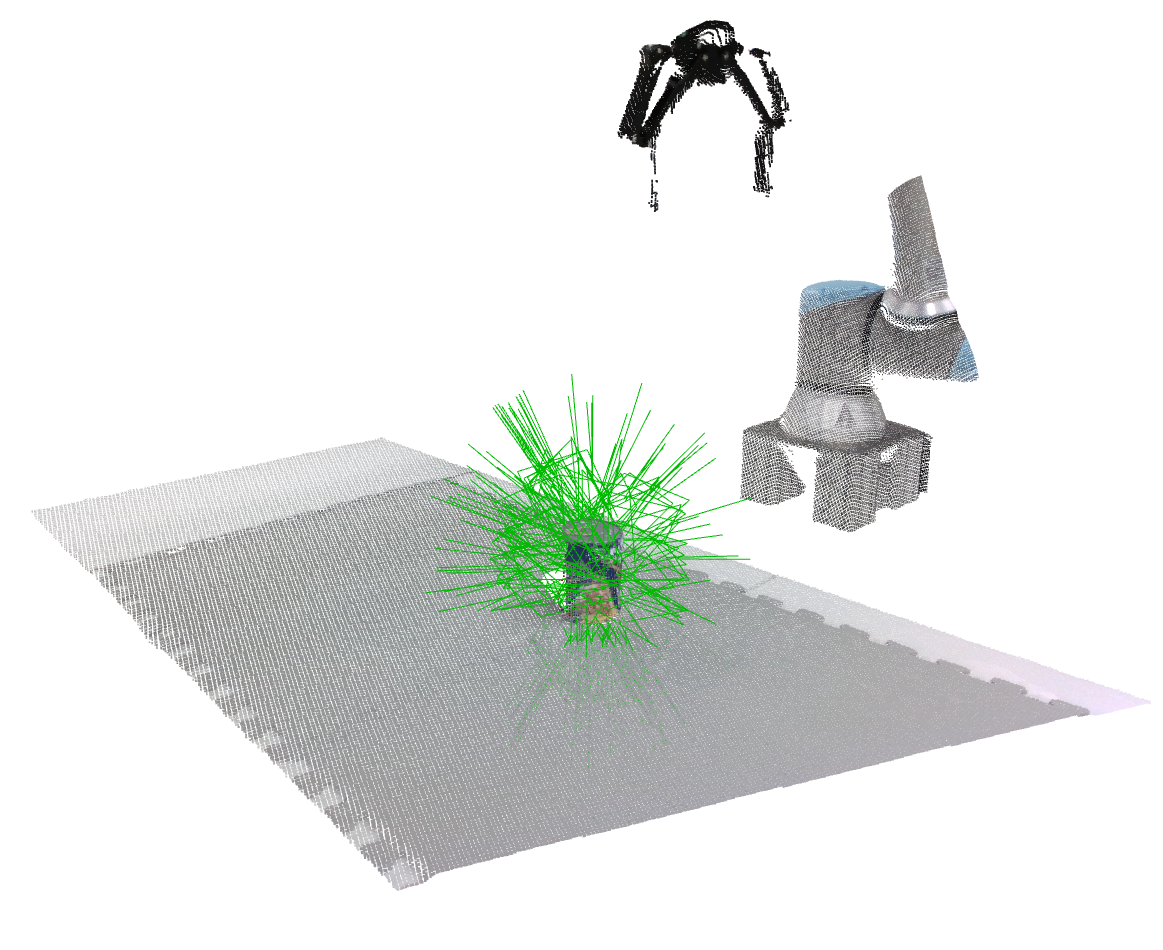} & \includegraphics[width=\linewidth]{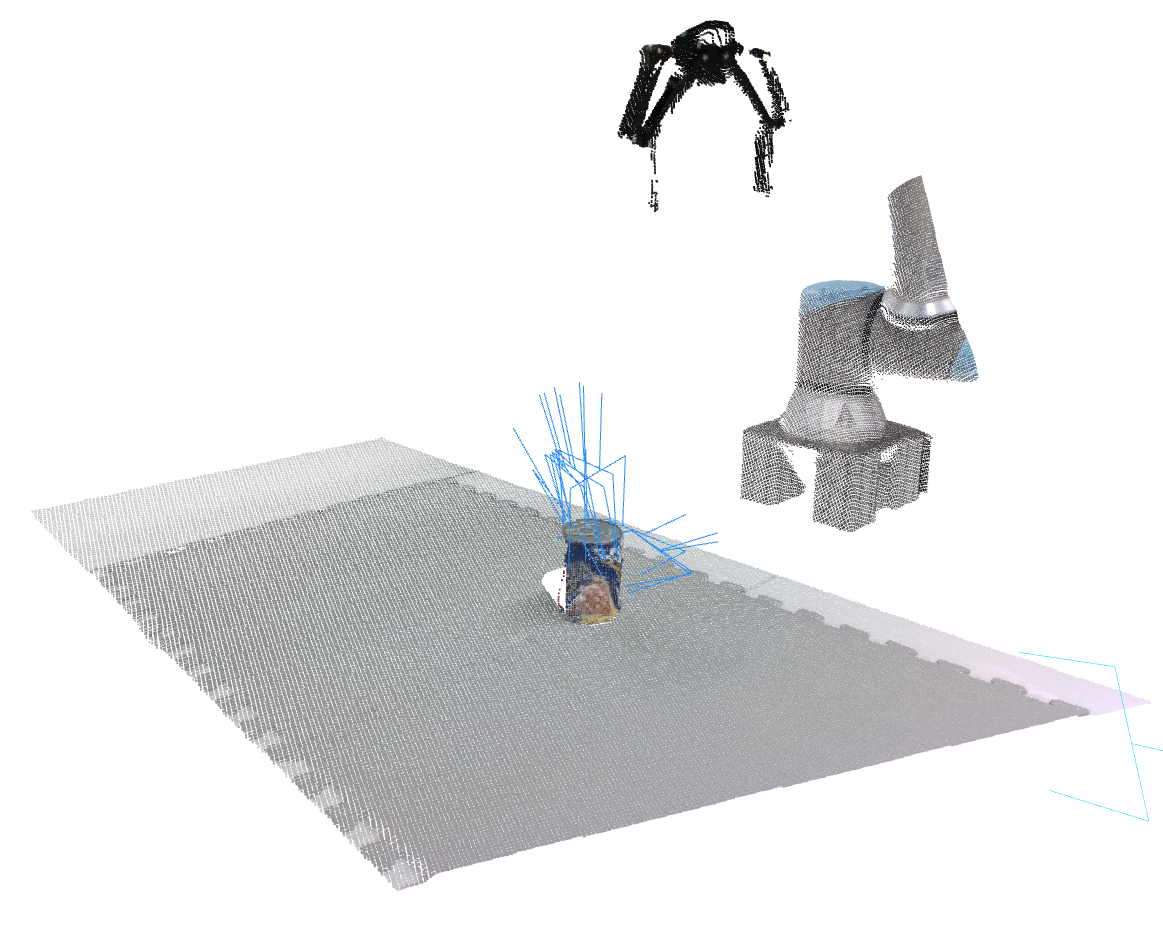}
        \\ \bottomrule
    \end{tabularx}
    \vspace{2pt}
    \caption{\small Qualitative grasp predictions from the real robot experiments overlaid on the colored point cloud in the robot frame, for \ourmodel (middle column) and the M2T2 \cite{yuan2023m2t2} baseline (right column). From top to bottom, we show a representative example from each of the environments (row 1-3 is in clutter): shelf, basket, tabletop, single isolated object. The target object to grasp is highlighted in \textcolor{red}{red} on the left column. \ourmodel only generates grasps for the target object~(\textcolor{myGreen}{green} grasps in middle column). Since M2T2 is a scene-centric model, we plot the predicted grasps for all objects in  \textcolor{myLightBlue}{light blue} and the predictions specific for the target object in \textcolor{myDarkBlue}{dark blue}. Overall, \ourmodel's predictions are more focused on the target object and have greater coverage than M2T2.}
    \vspace{-15pt}
    \label{tab:grasp_predictions}
\end{table}

\subsection{Quantitative Metrics in Radar Chart in Sec.~\ref{sec:introduction}}
\label{supp:radarchart}

The radar chart in Sec.~\ref{sec:introduction} shows the aggregate performance of \ourmodel and the baselines in various settings. For the object-centric simulation experiments, the key metric was Area Under Curve (AUC) of a Precision-Coverage curve, as displayed in Fig \ref{fig:baseline_comparison} and \ref{fig:robotiq} for the Franka and Robotiq-2F-140 grippers respectively. For the Suction gripper, we did not train a M2T2 model since we were not able to directly apply the contact-graspnet representation. Instead, we simply report the coverage metric comparing just the generators of \ourmodel and SE3-Diff \cite{urain2022se3dif}. For both the FetchBench and real robot experiments, we report grasp success rates shown in Fig \ref{fig:fetchbench} and Table \ref{tb:robot_v2} respectively.

\begin{table*}[]
\centering
\resizebox{\linewidth}{!}{
    \begin{tabular}{@{}cccccccc@{}}
    \toprule
    Dataset          & Year                & \#Grippers      & \#Objects & \#Grasps        & Grasp Label   & Synthesis Method        & Code + Data\\ \midrule
    HO-3D~\cite{hampali2020honnotate-ho3d} & 2020   & 1 (Human hand)             & 10                  & 78K                   & Only +ve               & Human Demo                      & \textcolor{green}{\cmark} \\

    EGAD~\cite{morrison2020egad}      & 2020        & 1 (2-finger)     & 2,331                & 233K                  & Only +ve               & Evolutionary Algorithm    & \textcolor{green}{\cmark} \\

    DDG~\cite{liu2020deep-ddg}     & 2020           & 1 (5-finger)           & 500                 & 50K                   & Only +ve   & GraspIt + modified Q1~\cite{ferrari1992planning-Q1} & \textcolor{red}{\xmark} \\

    DexYCB~\cite{chao2021dexycb}       & 2021       & 1 (Human hand)             & 20                  & 582K                  & Only +ve               & Human Demo                      & \textcolor{green}{\cmark}       \\
    
    Acronym ~\cite{eppner2021acronym}  & 2021       & 1 (2-finger)     & 8,872                & 17.7M                 & +ve \& -ve & Flex~\cite{macklin2014unified-flex-simulator} & \textcolor{green}{\cmark} \\
    UniGrasp ~\cite{shao2020unigrasp}  & 2020       & 12 (2 \& 3finger)     & 1000                &  2M+                   & Only +ve          & Contact Points Network + FastGrasp~\cite{pokorny2013classical-fastgrasp}                                 & \textcolor{green}{\cmark} \\
    
    DexGraspNet~\cite{wang2023dexgraspnet}  & 2023  & 1 (5-finger)           & 5,355                & 1.3M                  & Only +ve               & Differentiable grasping  & \textcolor{green}{\cmark} \\
    Fast-Grasp'D~\cite{turpin2023fast-graspd} & 2023 & 3 (3-5 finger)       & 2,350                & 1M                    & Only +ve               & Differentiable grasping  & \textcolor{red}{\xmark} \\
    GenDexGrasp~\cite{li2023gendexgrasp}  & 2023    & 5 (2-5 finger)       & 58                  & 436K                  & Only +ve               & Differentiable grasping  & \textcolor{green}{\cmark} \\
    MultiGripperGrasp       & 2024               & 11 (2-5 finger \& Human) & 345        & 30.4M         & Ranked & GraspIt + Isaac Sim~\cite{nvidia2023-isaac-sim} & \textcolor{green}{\cmark}      \\
    \midrule
    \textbf{GraspGen (Ours)}       & \textbf{2025}               & \textbf{3 (2-finger \& Suction)} & \textbf{8,515}        & \textbf{53.1M}         & \textbf{+ve \& -ve} & \textbf{Sampling + Isaac Sim~\cite{nvidia2023-isaac-sim}} & \textcolor{green}{\cmark}  \\ \bottomrule
\end{tabular}
}
\captionsetup{justification=centering}
\caption{Comparison of GraspGen with existing grasping datasets. +ve and -ve denote positive and negative grasp samples, respectively. Adapted from \cite{casas2024multigrippergraspdatasetroboticgrasping}.}
\label{tb:related-datasets}
\end{table*}

\subsection{Baseline Comparisons for Robotiq-2F-140}
\label{supp:robotiq}

\begin{figure}[h]
    \centering
    \captionsetup{justification=centering}
    \includegraphics[width=0.6\linewidth]{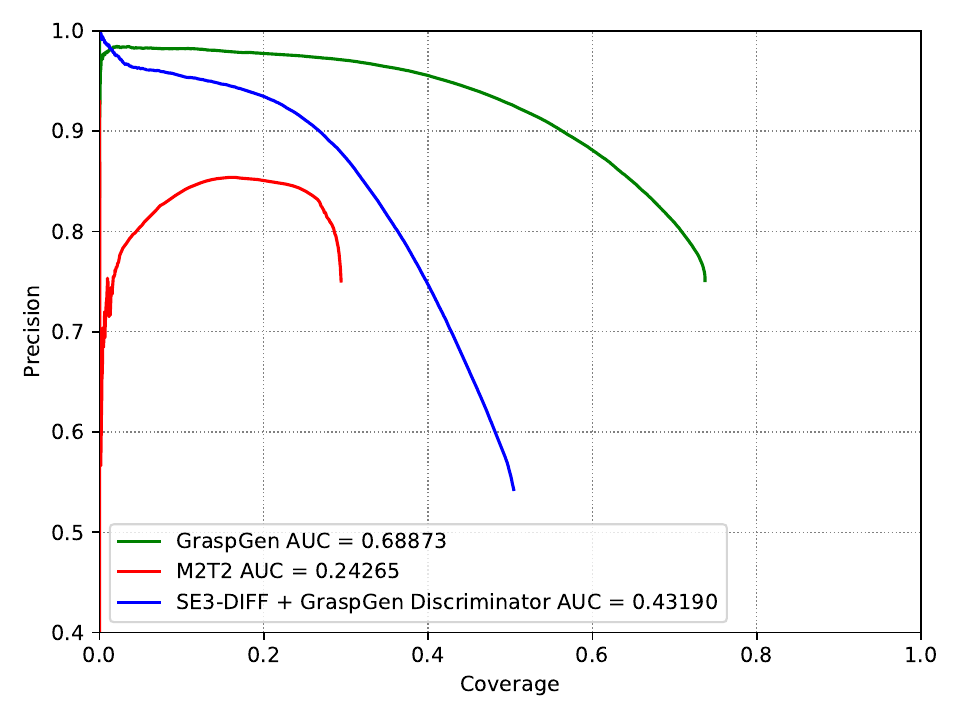}
    \caption{Baseline Comparisons for Robotiq-2F-140}
    \label{fig:robotiq}
\end{figure}

As summarized in Fig.~\ref{fig:robotiq}, \ourmodel outperforms both baselines by a substantial margin - almost double the AUC as M2T2~\cite{yuan2023m2t2}. M2T2 uses the contact point formulation from ~\cite{sundermeyer2021-contact-graspnet}, which is designed for symmetric, non-adaptive grippers pinch grippers and hence does not directly transfer to adaptive grippers like the Robotiq-2F-140.

Due to the poor performance of this Robotiq-trained M2T2~model, for real-world robot experiments~(see~Table~\ref{tb:robot_v2} in Sec.~\ref{sec:realrobot}) we re-use the M2T2 Franka Panda model with a fixed translation offset~(-10$cm$ along the $z$ axis).

\subsection{Baseline comparisons for Suction}
\label{supp:suction}
Quantitative comparisons of the \ourmodel generator is shown in Table \ref{tb:suction_results}. The rotation error is large for suction, since the grasps are symmetric along the approach direction. Surprisingly, \ourmodel achieves a slightly higher L2 translation error, even though its coverage is substantially larger than SE3-Diff.

In terms of learning difficulty, the embodiments rank from hardest to easiest as: Franka, suction, Robotiq.
We attribute this to gripper complexity -- suction is symmetric along the approach direction, while the Robotiq gripper is underactuated.

\subsection{Additional Ablations}
\label{supp:ablations}

\subsubsection{Ablation on Dataset}
We want to demonstrate that our proposed \ourmodel datasets are comparable with prior datasets. More specifically, we compare to the ACRONYM \cite{eppner2021acronym} dataset which is the most widely used 6-DOF grasping dataset for the Franka gripper used in \cite{mousavian2019-6dofgraspnet, sundermeyer2021-contact-graspnet, yuan2023m2t2, urain2022se3dif}. We trained two  models with the recipe shown in Algorithm \ref{alg:graspgen_training_recipie}, on both the ACRONYM and \ourmodel-Franka dataset, and tested separately on their corresponding test sets.

\begin{figure}[h]
    \centering
    \begin{subfigure}{0.48\textwidth}  
\captionsetup{justification=centering}
\includegraphics[width=\linewidth]{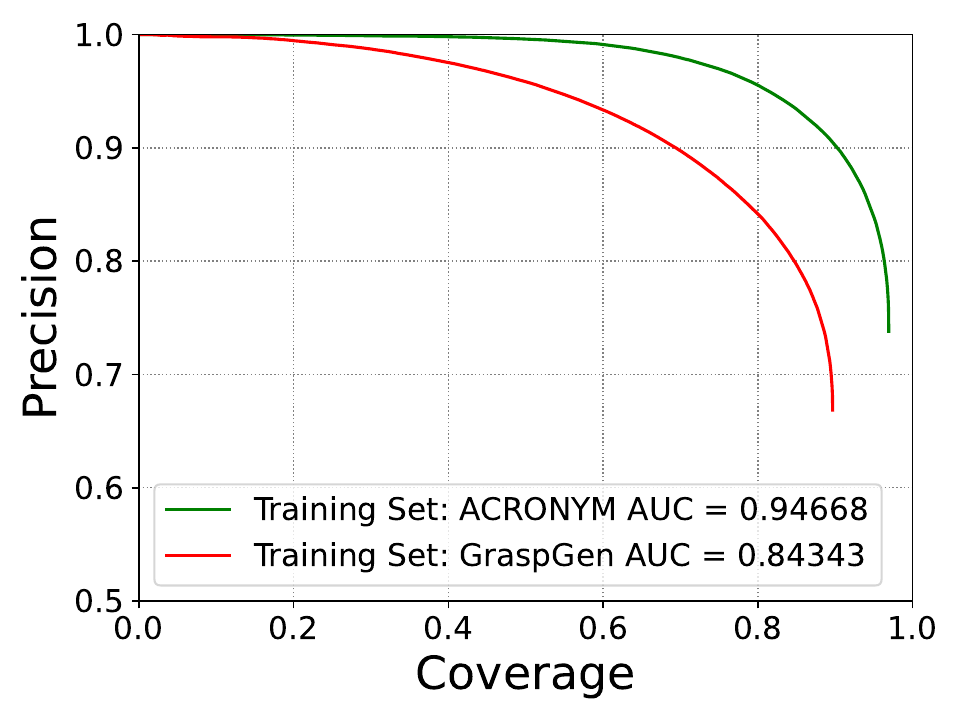}
        \caption{ACRONYM Franka test set \cite{eppner2021acronym}}
    \end{subfigure}
    \begin{subfigure}{0.48\textwidth}  

\captionsetup{justification=centering}
\includegraphics[width=\linewidth]{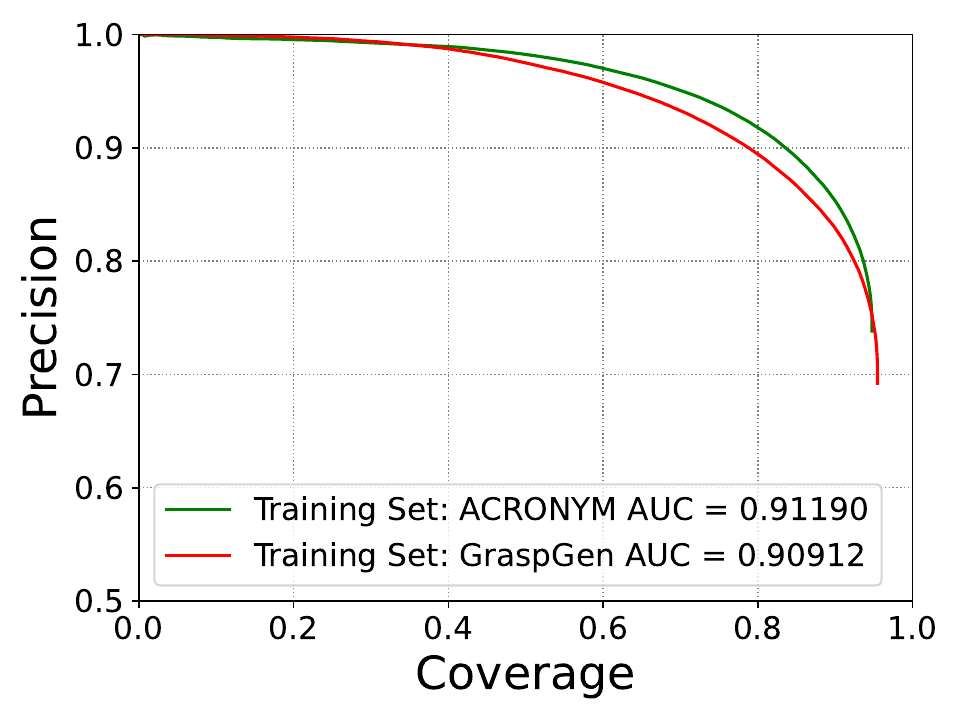}
        \caption{\ourmodel-Franka test set}
    \end{subfigure}
\captionsetup{justification=centering}
    \caption{Dataset Ablation}
    \label{fig:dataset_ablation}
\end{figure}

Both models achieved a overall proficient performance, though the model trained with ACRONYM is slightly better than GraspGen-Franka, including on the test set of the latter. We hypothesize that this is due to a mismatch in the simulator (ACRONYM was generated with Flex physics engine while we used Physx) and shape datasets (ACRONYM used ShapeNet~\cite{savva2015semantically} while \ourmodel uses Objaverse~\cite{deitke2023objaverse}).

\subsubsection{Analysis of the Discriminator}
\label{appendix:discriminator}
\begin{figure}[h]
    \captionsetup{justification=centering}
    \centering
    \includegraphics[width=0.6\linewidth]{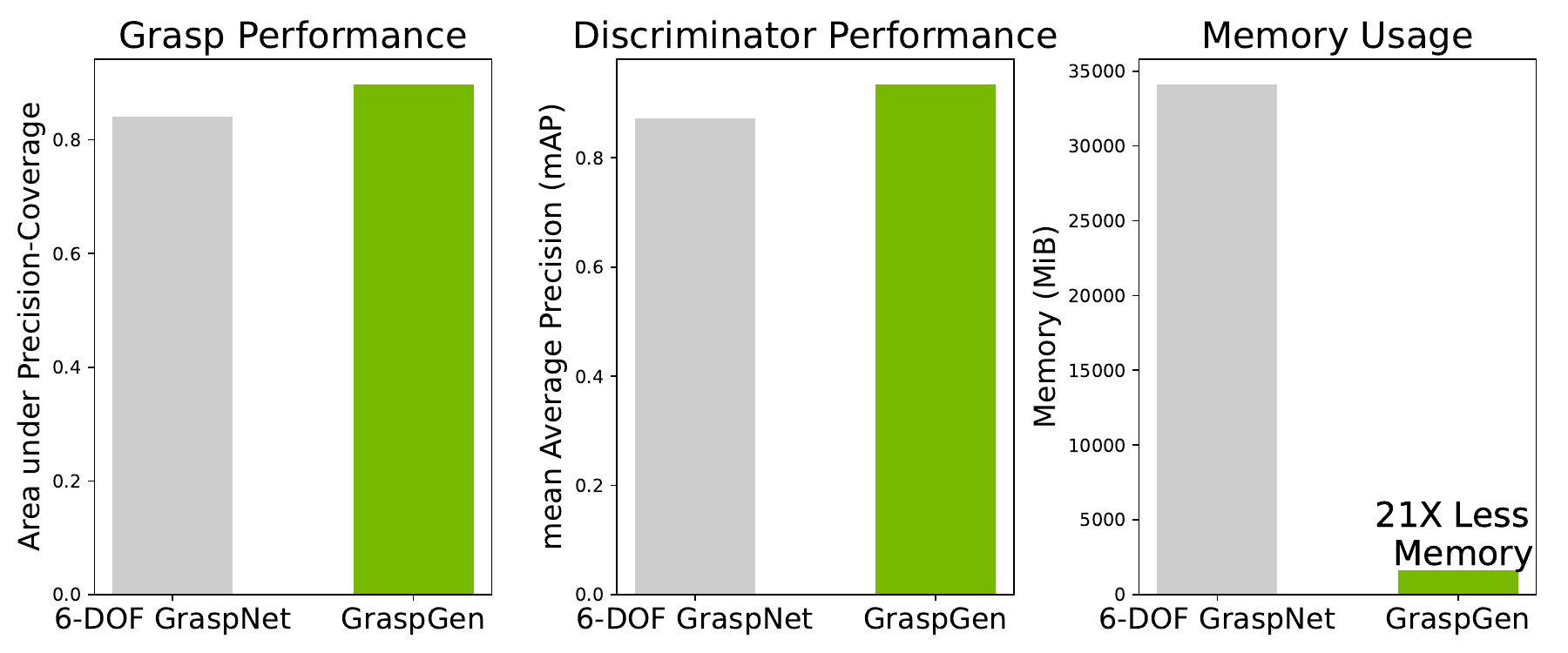}
    \caption{Discriminator of \ourmodel vs. 6-DOF GraspNet \cite{mousavian2019-6dofgraspnet}}
    \label{fig:system_compare}
\end{figure}

As highlighted in~\cite{lum2024get}, the progress in grasp discriminator has been much less than in grasp generation. As such, we compare to the discriminator architecture proposed in \cite{mousavian2019-6dofgraspnet}, as shown in Fig.~\ref{fig:system_compare}. We observe that our discriminator is more performant in terms of accuracy metrics (6.7$\%$ and 5.87$\%$ higher in AUC and mean Average Precision (mAP) of the binary classification sigmoid scores) and uses 21$\times$ less memory for the same batch size compared to~\cite{mousavian2019-6dofgraspnet}.

This is because in \cite{mousavian2019-6dofgraspnet}, the grasp poses $g \in SE(3)$ were transformed to a grasp point cloud $X \in \mathbb{R}^{N\times3}$ (where $N=5$ are predefined set of points on the gripper) and input to a PointNet++~\cite{pointnet} backbone with an additional segmentation label to specify the points from the gripper vs. object. This caused the GPU memory to scale $\mathcal{O}(N \times B)$ where $B$ is the batch size. However, in \ourmodel as shown in Fig.~\ref{fig:network}, we simply pass in the grasp poses in $SE(3)$ (i.e. memory scales with $\mathcal{O}(B)$ instead) without any point cloud duplication. We also re-use the object encoder (the biggest part of the network) weights without retraining in the generator for the discriminator, resulting in significantly smaller network than~\cite{mousavian2019-6dofgraspnet}.

\begin{table}
\footnotesize
\centering
\captionsetup{justification=centering}
\caption{Ablation on Rotation Representation}
\label{tb:ablation_gen_rotation}
\resizebox{0.48\textwidth}{!}{%
    \begin{tabular}{c|c|ccc}
    \toprule
      \begin{tabular}{@{}c@{}} \textbf{Represen-}\\ \textbf{tation}  \end{tabular} & \textbf{Limits} &  \begin{tabular}{@{}c@{}} \textbf{Translation} \\ \textbf{Error (cm)} \end{tabular}  & \begin{tabular}{@{}c@{}} \textbf{Rotation} \\ \textbf{Error (rad)} \end{tabular} & \begin{tabular}{@{}c@{}} \textbf{Coverage} \\ \textbf{(\%)} \end{tabular}    \\
    \midrule
    6D \cite{Zhou_2019_CVPR} & [-1, 1] & 
    \begin{tabular}{@{}c@{}} 3.126 \\ ($\pm$ 0.0372) \end{tabular} & 
    \begin{tabular}{@{}c@{}} 0.557 \\ ($\pm$ 0.003) \end{tabular} & 
    \begin{tabular}{@{}c@{}} 84.86 \\ ($\pm$ 0.301) \end{tabular} \\ \midrule
    Euler angles & [-$\pi$, $\pi$] & 
    \begin{tabular}{@{}c@{}} 3.047 \\ ($\pm$ 0.0045) \end{tabular} & 
    \begin{tabular}{@{}c@{}} 0.541 \\ ($\pm$ 0.001) \end{tabular} & 
    \begin{tabular}{@{}c@{}} 84.78 \\ ($\pm$ 0.142) \end{tabular} \\ \midrule
    Lie Algebra \cite{urain2022se3dif} & [-$\pi$, $\pi$] &  
    \begin{tabular}{@{}c@{}} \textbf{3.008} \\ ($\pm$ \textbf{0.0200}) \end{tabular} & 
    \begin{tabular}{@{}c@{}} \textbf{0.535} \\ ($\pm$ \textbf{0.002}) \end{tabular} & 
    \begin{tabular}{@{}c@{}} \textbf{84.86} \\ ($\pm$ \textbf{0.096}) \end{tabular} \\
    \bottomrule
    \end{tabular}
}
\end{table}

\subsubsection{Ablation on Rotation representation}
\label{appendix:rotation}
We compare three popular rotation representations in grasp learning and computer vision, summarized in Table~\ref{tb:ablation_gen_rotation}. All input values are scaled to [-1,1] before being passed into the diffusion model. The 6D rotation representation, which concatenates the first two columns of a rotation matrix, is widely used in computer vision \cite{Zhou2018}. We found that these two, along with Euler angles, performed comparably. This suggests that proper normalization is the key factor for effective diffusion model learning on large grasp datasets. For all other experiments, we use the Lie algebra representation.

\begin{table}[]
\footnotesize
\centering
\captionsetup{justification=centering}
\caption{Ablation on Generator Backbone}
\label{tb:ablation_gen_backbone}
\resizebox{0.48\textwidth}{!}{%
    \begin{tabular}{c|ccc}
    \toprule
     \begin{tabular}{@{}c@{}} \textbf{Object} \\ \textbf{Encoder} \end{tabular} & \begin{tabular}{@{}c@{}} \textbf{Translation} \\ \textbf{Error (cm)} \end{tabular}  & \begin{tabular}{@{}c@{}} \textbf{Rotation} \\ \textbf{Error (rad)} \end{tabular} & \begin{tabular}{@{}c@{}}\textbf{Coverage} \\ \textbf{(\%)} \end{tabular}   \\
    \midrule
    PointNet++ \cite{pointnet} & 
    \begin{tabular}[c]{@{}c@{}}3.724 \\ ($\pm$ 0.0221)\end{tabular} & 
    \begin{tabular}[c]{@{}c@{}}0.637 \\ ($\pm$ 0.0011)\end{tabular} & 
    \begin{tabular}[c]{@{}c@{}}79.15 \\ ($\pm$ 0.114)\end{tabular} \\ \midrule
    PointTransformerV3 \cite{wu2024ptv3} & 
    \begin{tabular}[c]{@{}c@{}}\textbf{3.126} \\ ($\pm$ \textbf{0.0372})\end{tabular} & 
    \begin{tabular}[c]{@{}c@{}}\textbf{0.557} \\ ($\pm$ \textbf{0.003})\end{tabular} & 
    \begin{tabular}[c]{@{}c@{}}\textbf{84.86} \\ ($\pm$ \textbf{0.301})\end{tabular} \\ \bottomrule
    \end{tabular}
}
\end{table}

\subsubsection{Ablation on Pointcloud Encoder}
\label{appendix:pointcloudencoder}
PointNet++ \cite{pointnet} remains the most widely used backbone for encoding point clouds in robotics. While transformer-based architectures have advanced significantly, their adoption in robotics remains limited.
We demonstrate substantial gains using the SOTA transformer backbone PointTransformerV3~(PTv3)~\cite{wu2024ptv3}. As shown in Table~\ref{tb:ablation_gen_backbone}, PTv3 reduces translation error by~\SI{5.3}{mm} and increases recall by~\SI{4}{\percent}. We hypothesize that this performance gap will further widen with larger-scale data.

\subsection{Further Details of FetchBench Experiments}
\label{supp:fetchbench}
We reimplement FetchBench~\cite{han2024fetchbench} in Isaac Sim \cite{nvidia2023-isaac-sim}. Since we want to only compare the performance of grasp generation methods, agnostic of the motion planners, we use the ground truth collision mesh of the scene. We specifically use the ``FetchMeshCurobo" (for the oracle planner) and ``FetchMeshCuroboPtdCGNBeta" (for Contact-GraspNet \cite{sundermeyer2021-contact-graspnet}, which only makes one attempt at grasping without any re-trials.

\subsection{Computing Earth-Movers-Distance (EMD) in Fig.~\ref{fig:dataset_emd_and_ongenerator_ablation}}
\label{supp:emd}

We now describe how we computed the Earth-Movers-Distance for Fig \ref{fig:dataset_emd_and_ongenerator_ablation}. We want to measure the distribution shift between the data generated by the diffusion model $\mathcal{G}$ compared to the training dataset $\hat{\mathcal{G}}$. Given these two datasets for the same object, we subsample 500 grasps from each. For each pose $g_{i} \in \mathcal{G}$ and $g_{j} \in \hat{\mathcal{G}}$, we measure the pair-wise distance using the cost function introduced in \cite{urain2022se3dif}:

\begin{equation}
    d(g_i, g_j) = \| t_i - t_j \| + \left\| \mathrm{LogMap}(\mathbf{R}_i^{-1} \mathbf{R}_j) \right\|
\end{equation}

We then solve a Linear Sum Assignment optimization problem, which effectively searches for the one-to-one assignment between the samples in both distributions based on the lowest distance. We repeat this process for 5 random subsamples of 500 grasps from both distributions, and average to get a score for each object.

\subsection{Data Augmentation for Sim2Real transfer}
\label{supp:augmentation}

We found that applying noise and data augmentations were crucial for sim2real transfer. We apply the following randomizations to the point clouds at every training iteration:
\begin{itemize}
    \item Randomized orientation after point cloud mean centering
    \item Random camera viewpoints
    \item Random subsampled sets of points
    \item Instance segmentation error
\end{itemize}

\begin{figure}[h]
    \centering
    \includegraphics[width=0.6\linewidth]{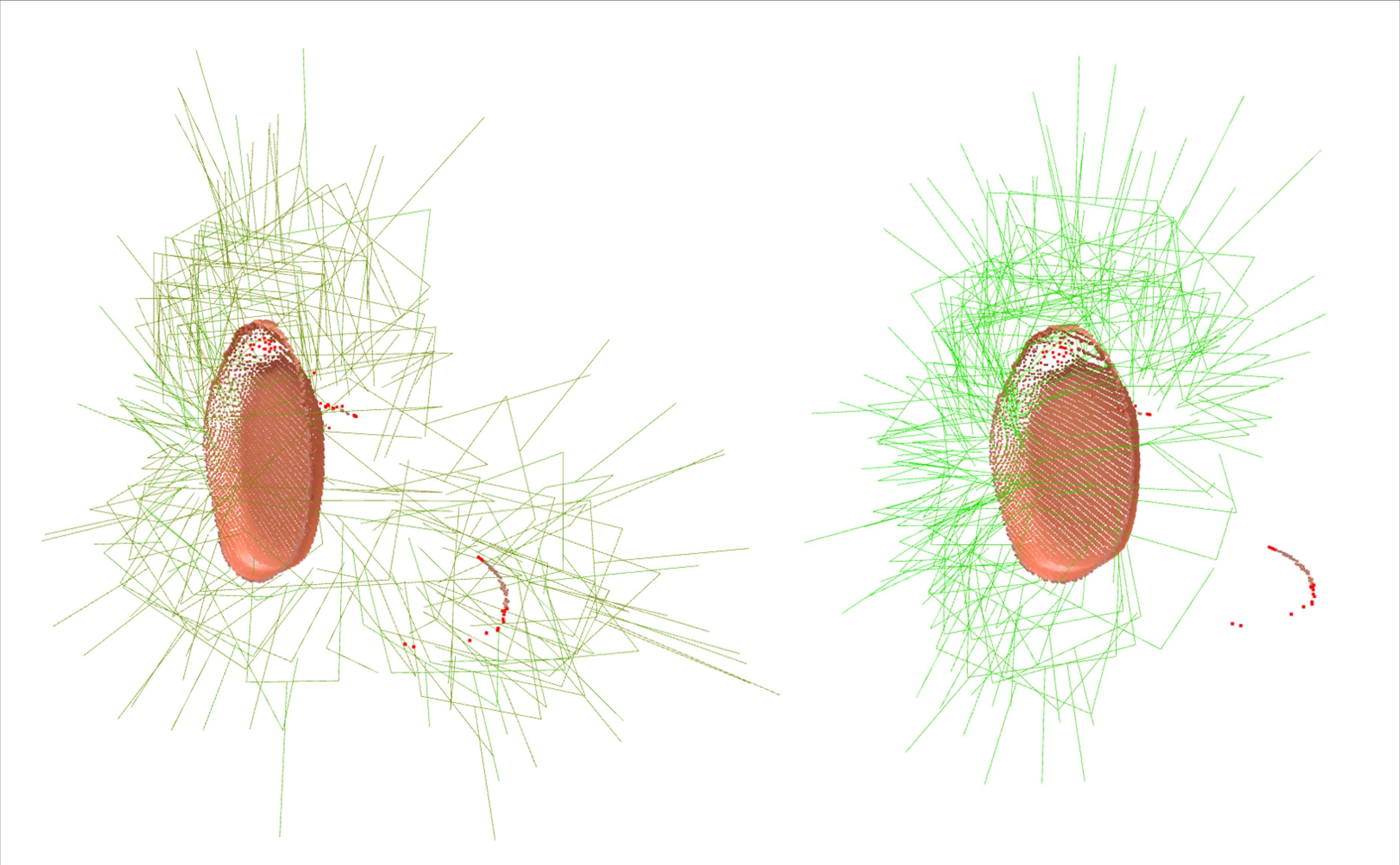}
    \caption{Grasp model predictions without (left) and with (right) instance segmentation noise augmentation. Notice how the outlier points on the bottom right are ignored in the latter model.}
    \label{fig:augmentation}
\end{figure}

While modern instance segmentation methods like SAM2\cite{ravi2024sam2} are very proficient, they sometimes suffer from overshooting pixels at object boundaries, leading to sizable geometric outliers when projected to 3D. As shown in Fig \ref{fig:augmentation} on the left (featuring an upright, orange plate), this causes the grasp network to predict grasps on the outlier regions with high confidence. These grasps are potentially unsafe, causing collisions between the robot and the table or walls. To train the model to be robust to such errors, we simulate instance segmentation error during training. We use Scene Synthesizer \cite{Eppner2024} to place objects on support surfaces, render the object segmentation mask and dilate them to create artificial outliers. Our model trained on such augmentations is robust to such errors as shown in Fig \ref{fig:augmentation} on the right.

\subsection{Inference Parameter Tuning}
\label{supp:tuning}

\begin{figure}[h]
    \centering
    \captionsetup{justification=centering}
    \includegraphics[width=0.6\linewidth]{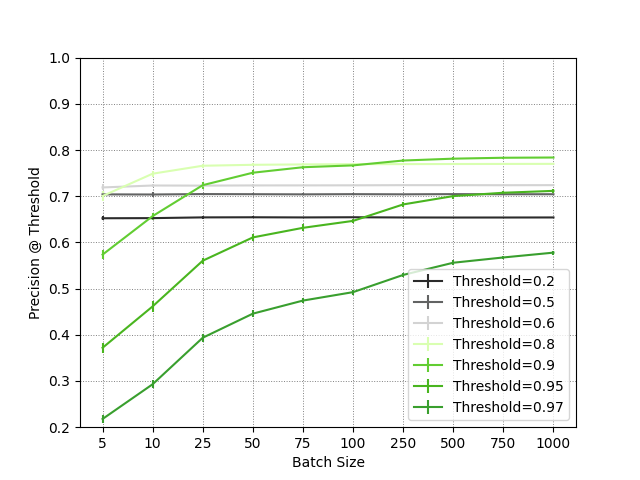}
    \caption{Test-time tuning of Batch Size}
    \label{fig:inference_parameter_tuning}
\end{figure}

After a \ourmodel model is trained, there are two key hyperparameters that a user has to select at inference time: (1) the threshold to filter out grasps of lower quality as well as (2) the number of grasps sampled through the diffusion model.
In Fig.~\ref{fig:inference_parameter_tuning} we investigate the relationship between the batch size sampled (horizontal axis) and the threshold with the final success rate/precision of the grasp set filtered by the said threshold. If the threshold is set very low (below 0.5), the precision of the grasps suffers as expected. However, as the threshold is increased, the number of grasps remaining after thresholding also reduces and lowers the the precision. When setting a high threshold, one would need to sample a large batch size for best performance.

\begin{table}
\footnotesize
\centering
\captionsetup{justification=centering}
\caption{Baseline Comparisons for Suction}
\label{tb:suction_results}
\begin{tabular}{c|ccc}
\hline
 \begin{tabular}{@{}c@{}} \textbf{Object} \\ \textbf{Encoder} \end{tabular} & \begin{tabular}{@{}c@{}} \textbf{Translation} \\ \textbf{Error (cm)} \end{tabular}  & \begin{tabular}{@{}c@{}} \textbf{Rotation} \\ \textbf{Error (radians)} \end{tabular} & \begin{tabular}{@{}c@{}}\textbf{Coverage} \\ \textbf{(\%)} \end{tabular}   \\
\midrule
\ourmodel & 7.79 & 1.83 & 73.1 \\ \midrule
SE3-Diff \cite{urain2022se3dif} & 6.12 & 1.87 & 38.5 \\ \hline
\end{tabular}
\end{table}

\end{document}